%% file: main-arxiv.tex
\PassOptionsToPackage{table,dvipsnames}{xcolor}
\documentclass[10pt,twocolumn,letterpaper]{article}
\input{preamble-arxiv}
\title{\method{}: Bridging Slow Vision and Fast Touch \\ with a World–Language–Action Model}
\author{
Kaichen Zhou\textsuperscript{1,2,*}\quad
Yuzhen Chen\textsuperscript{2,*}\quad
Hannah Yao\textsuperscript{3,*}\quad
Chonghe Jiang\textsuperscript{1}\\
Rao Fu\textsuperscript{4}\quad
Xihang Yu\textsuperscript{1}\quad
Wei Dai\textsuperscript{1}\quad
Bashar Kabbarah\textsuperscript{1}\\
Haoyu Chen\textsuperscript{2}\quad
Ishita Goluguri\textsuperscript{1}\quad
Zhiyang Dou\textsuperscript{1}\quad
Yue Ma\textsuperscript{5}\\
Haonan Chen\textsuperscript{2}\quad
Enze Xie\textsuperscript{6}\quad
Fangneng Zhan\textsuperscript{1}\quad
Yilun Du\textsuperscript{2}\\
Mengyu Wang\textsuperscript{2}\quad
Wojciech Matusik\textsuperscript{1}\quad
Antonio Torralba\textsuperscript{1}\quad
Paul Pu Liang\textsuperscript{1}\\[0.6em]
{\small
\textsuperscript{1}Massachusetts Institute of Technology\quad
\textsuperscript{2}Harvard University}\\
{\small
\textsuperscript{3}University of California, Davis\quad
\textsuperscript{4}University of California, Berkeley}\\
{\small
\textsuperscript{5}Stanford University\quad
\textsuperscript{6}NVIDIA}\\[0.3em]
{\small \textsuperscript{*}Equal contribution.}
}
\begin{document}
\makeatletter
\let\originalmaketitle\@maketitle
\def\@maketitle{%
  \originalmaketitle
  \input{section/00_teaser}}
\makeatother
\maketitle
\input{section/00_abstract}
\input{section/01_introduction}
\input{section/02_related_work}
\input{section/03_methodology}

\input{section/06_evaluation}

\input{section/07_conclusion}
\input{section/ai_disclosure}
\bibliographystyle{ieeenat_fullname}
\bibliography{references}

\clearpage
\label{sec:supplement-start}

\appendix
\startcontents[appendix]
\section*{Appendix}
\makeatletter
\def\@currentlabel{Appendix}
\makeatother
\label{sec:appendix}
\setcounter{tocdepth}{2}
\printcontents[appendix]{}{1}{}
\input{section/09_appendix}

\clearpage

\end{document}

%% file: preamble-arxiv.tex
\usepackage{cvpr}
\usepackage[T1]{fontenc}
\usepackage{graphicx,amsmath,amssymb,amsfonts,bm,booktabs,array,multirow,makecell}
\usepackage{xcolor,amsthm,algorithm,algpseudocode,xr}
\definecolor{cvprblue}{rgb}{0.21,0.49,0.74}
\usepackage{annotate-equations}
\usepackage{placeins}
\usepackage[pagebackref,breaklinks,colorlinks,allcolors=cvprblue]{hyperref}
\def\confName{CVPR}
\def\confYear{2027}
\newcommand{\method}{T-WLA}

\definecolor{col1}{RGB}{232,161,148}
\definecolor{col2}{RGB}{148,187,232}

\hypersetup{pdftitle={T-WLA: Bridging Slow Vision and Fast Touch},pdfauthor={Anonymous Authors}}
\newcommand{\V}{\mathrm{V}}
\newcommand{\T}{\mathrm{T}}
\newcommand{\inp}{\mathrm{in}}
\newcommand{\base}{\mathrm{base}}
\newcommand{\vis}{\mathrm{vis}}

\newcommand{\act}{\mathrm{act}}

\providecommand{\eqnannotationfont}{}
\definecolor{NavyBlue}{RGB}{31,78,121}
\definecolor{OliveGreen}{RGB}{85,107,47}
\definecolor{BurntOrange}{RGB}{204,85,0}
\definecolor{Plum}{RGB}{142,69,133}
\definecolor{WildStrawberry}{RGB}{214,54,100}

\usepackage{titletoc}

%% file: section/00_teaser.tex
\begingroup
  \centering

  \includegraphics[width=1.0\linewidth]{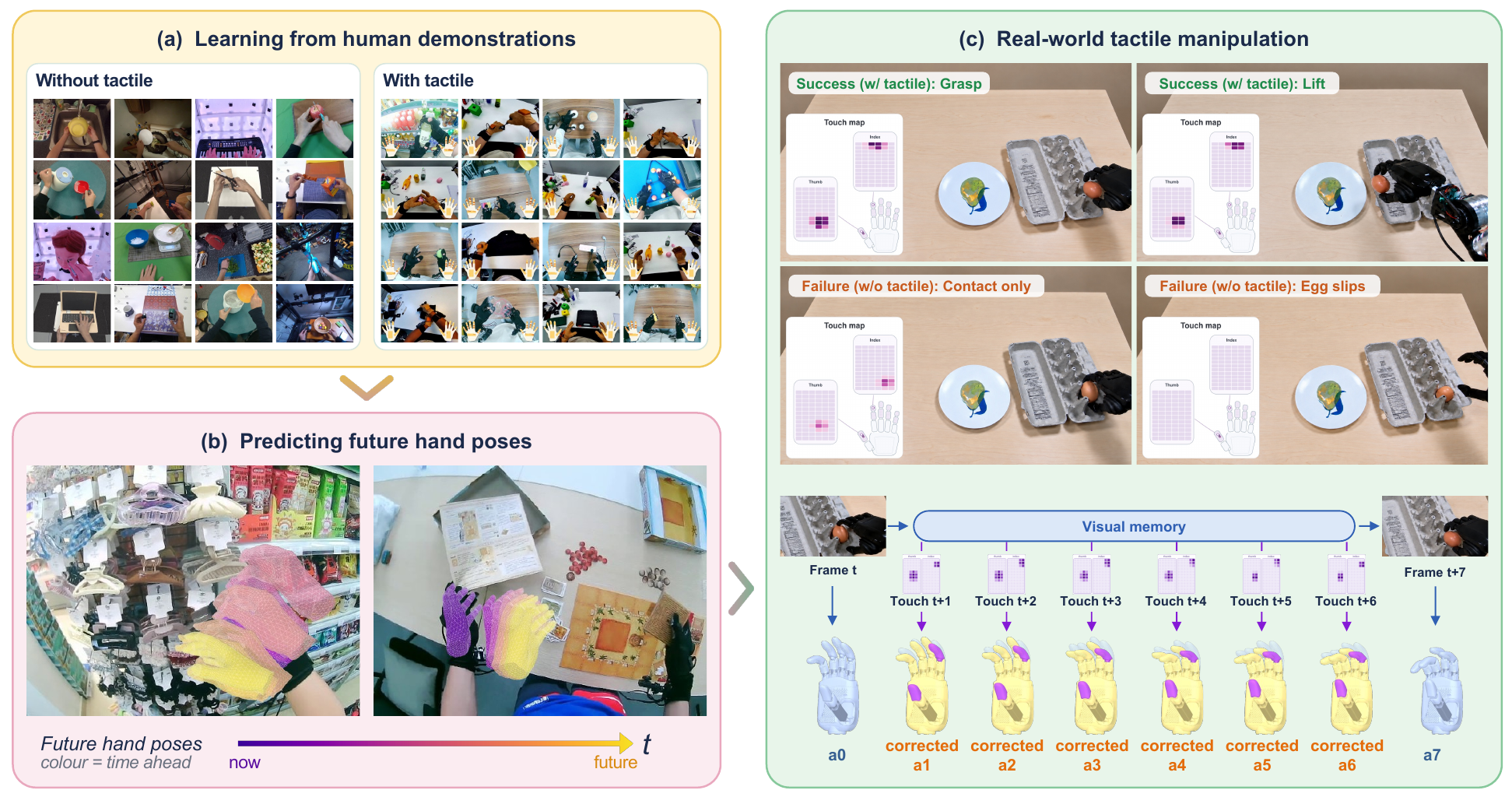}
  \par\captionof{figure}{
\textbf{Tactile World--Language--Action (\method{}).}
The visual branch uses current and past visual information to produce an action chunk and visual memory.
Between visual updates, the tactile editor uses this cached visual memory and recent touch to correct the action chunk.
During training, predicting future images and pressure helps the model learn features useful for control.
}
\vspace{0.5cm}
  \label{fig:overview}
\par\endgroup

%% file: section/00_abstract.tex
\begin{abstract}
In contact-rich manipulation, contact conditions can change before a visual policy updates its action plan. 
Recomputing the plan through the visual policy adds latency, while tactile feedback alone does not specify how the ongoing motion should be revised for the task. 
We introduce \method{}, a world--language--action architecture that uses shared visual memory to guide both action generation and tactile correction.
At each visual update, the visual branch integrates the current image, task instruction, and retrieved history into a visual memory that conditions base-action generation.
The same memory is cached for a lightweight tactile editor, which combines it with the active action chunk and incoming touch to predict residual corrections to unexecuted actions without rerunning the visual branch.
Future-image and future-pressure prediction provide training-only supervision, encouraging the visual memory and tactile features to capture interaction dynamics relevant to action generation and correction.
We evaluate \method{} on tactile-conditioned hand-motion prediction using EgoTouch and Tachin, and on four real-world dexterous manipulation tasks.
On EgoTouch, T-WLA reduces finger-joint MPJPE by 30\% relative to the strongest evaluated baseline. Across four real-robot tasks, T-WLA achieves 51.0\% mean success versus 23.0\% for the baseline, a gain of 28.0 percentage points.
Videos and code are available at the
\href{https://t-wla.github.io/}{project website}.
\end{abstract}

%% file: section/01_introduction.tex
\section{Introduction}
\label{sec:introduction}

Contact-rich manipulation refers to tasks whose success depends on precisely managing physical contact with objects or the surrounding environment~\citep{hogan2022contact,lee2019making}.
Tactile feedback complements vision by revealing local contact conditions that are difficult to infer from images alone \citep{niu2026t,song2025opentouch,xue2025reactive,zhou2026touchworld}.
For example, when inserting a plug into a socket, vision guides alignment and approach, while touch provides feedback on contact and insertion resistance.
Because these conditions can change between visual updates, tactile feedback supports rapid corrections to the ongoing motion.

Despite the importance of tactile feedback in contact-rich manipulation, many general-purpose robot policies rely primarily on vision. 
Vision--language--action (VLA) models adapt pretrained vision--language backbones to predict robot actions from visual observations and language instructions \citep{kim2024openvla,intelligence2025pi_}.
World--action models further integrate future-state prediction with action generation, linking control decisions to the anticipated evolution of the environment \citep{rift2026wam,yuan2026fastwam}.
Without tactile feedback, these models must infer local contact conditions from visual cues, which may limit their response to visually subtle contact changes during precise manipulation.

Recent work incorporates tactile feedback into robot control through variable-rate expert architectures \citep{niu2026t}, residual adaptation \citep{wang2026vitar}, and inference-time policy steering \citep{wu2026inference}. We investigate a complementary question: \emph{how can fast tactile feedback correct a slow visual policy between visual updates?} 
Our central idea is to condition asynchronous tactile corrections on persistent visual memory that retains task and scene information, enabling local contact feedback to guide adjustments within the ongoing visual plan.


We instantiate this idea in \method{}, a multi-rate architecture for asynchronous visual--tactile control
(Fig.~\ref{fig:overview}).
The visual branch generates an action chunk and a history-aware visual representation, which is cached until the next visual update.
Between these updates, a lightweight tactile editor combines the cached representation with recent tactile observations to produce high-frequency residual corrections to the ongoing action chunk.
This separation allows the controller to respond to changing contact conditions without repeatedly running the visual branch.
During training, future-image and future-pressure prediction regularize the control representations, encouraging them to capture information relevant to subsequent interactions.
Our contributions are summarized as follows:

\begin{itemize}

\item \textbf{Memory-guided asynchronous control.}
We propose \method{}, which uses the same history-aware visual memory to guide base-action generation and tactile correction.
A lightweight tactile editor uses this memory and the active action chunk to correct unexecuted actions between visual updates, without rerunning the visual branch.

\item \textbf{Predictive supervision for control.}
We regularize the visual memory through future-image prediction and the tactile features through both future-image and future-pressure prediction.
These auxiliary objectives complement action supervision, and their prediction heads are removed at inference.

\item \textbf{Evaluation across hand-motion prediction and robot manipulation.}
We show that \method{} can be adapted to four contact-rich robot tasks with only 30 teleoperated demonstrations per task, achieving 51.0\% average success compared with 23.0\% for T-ReX. These robot results complement a 30\% reduction in finger-joint MPJPE on EgoTouch relative to the strongest reported baseline, demonstrating its applicability to both human hand-motion prediction and dexterous robot control.

\end{itemize}

%% file: section/02_related_work.tex
\section{Related work}
\label{sec:related}

\subsection{Memory and Action Chunk Execution}
\label{sec:related-memory}

A single observation may not provide enough information to determine the next action. Memory-based policies address this limitation by retaining information from earlier observations; for example, MemoryVLA stores consolidated perceptual and cognitive entries that can be queried over long horizons
\citep{shi2026memoryvla}. A complementary line of work predicts temporally extended action sequences. ACT introduced transformer-based action chunking to reduce the effective decision horizon \citep{zhao2023act}, while generative policies such as Diffusion Policy, RDT-1B, and $\pi_0$ generate multi-step actions with diffusion or flow-based objectives \citep{chi2025diffusion,liu2025rdt,intelligence2025pi_}. Such temporal coherence can reduce responsiveness when new observations arrive during execution \citep{liu2025bidirectional,zeng2026revisiting}. 
\begin{figure*}[!t]
\centering
\includegraphics[width=0.92\linewidth]{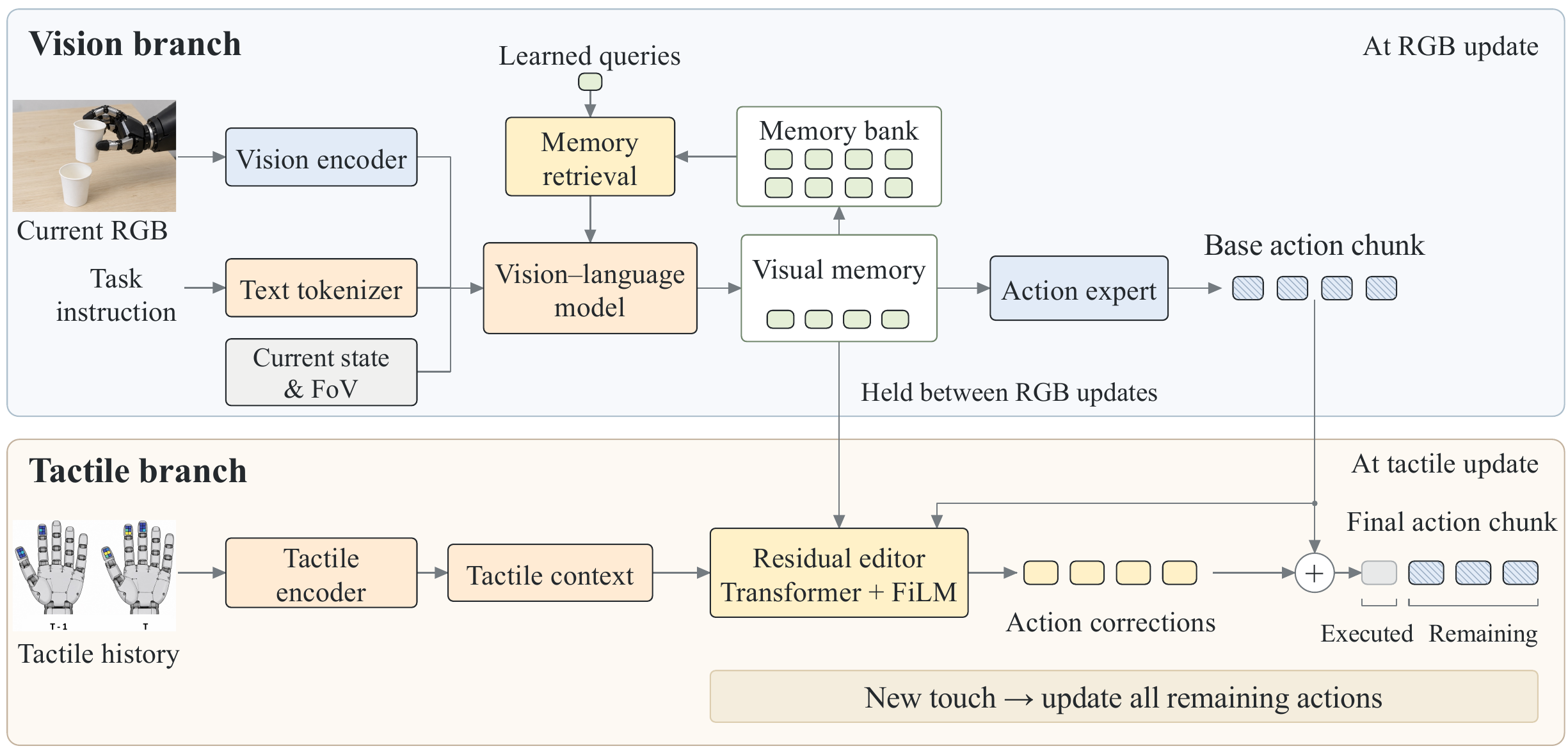}
    \vspace{-0.3cm}
\caption{
\textbf{Shared memory connects slow visual planning with fast tactile correction.}
Each visual update generates a base action chunk and visual memory.
Between updates, the tactile editor reads this cached pair and revises unexecuted actions.}
    \vspace{-0.3cm}
\label{fig:architecture}
\end{figure*}
Bidirectional Decoding improves closed-loop adaptation by selecting among candidate chunks
\citep{liu2025bidirectional}, while Consistency Policy and Streaming Diffusion Policy accelerate generative action synthesis \citep{prasad2024consistency,hoeg2025streaming}. Real-time chunking instead explicitly accounts for inference latency when transitioning between successive chunks \citep{black2026real}. \method{} addresses a complementary setting: it retains visual history across policy updates and uses high-rate tactile feedback to revise the unexecuted portion of an active action chunk.

\subsection{Tactile Foundation Models}
\label{sec:related-tactile}

Tactile representation learning aims to produce features that transfer across sensors and tasks \citep{bi2026vla, tao2026foretac, heng2025vitacformer, zhang2026tacvla, huang2025tactile,ma2026tacpac,chen2026pace,zhang2026retouch,zhou2026tacforcing,xue2026hitac,tian2026vt,higuera2026visuo}.
Heterogeneous Tactile Transformer learns a shared model across sensor types \citep{bi2026heterogeneous}, while FG-CLTP aligns tactile features with visual and language representations \citep{ma2026fg}.
Other work explores transferring representations learned from human tactile recordings to dexterous robots \citep{zhang2026human} and scaling pretraining through tactile foundation models \citep{team2026n0,zhou2026touchworld}.
Our tactile encoder follows this direction through contrastive pretraining on synchronized tactile and hand-pose histories.
Building on tactile representations, several systems study how to incorporate touch into manipulation policies.
Multi-rate architectures allow tactile feedback to influence control between visual updates.
T-Rex assigns vision and touch to experts operating at different rates \citep{niu2026t}, while Reactive Diffusion Policy combines a slow visual diffusion policy with a fast latent tactile controller \citep{xue2025reactive}.
FA-RDP adapts this division to task demand \citep{zhuo2026fa}, and DSWAM organizes world--action modeling into cooperating slow and fast systems \citep{zhu2026dswam}.
Residual adaptation provides another mechanism for incorporating touch into existing policies.
ViTaR attaches a lightweight visuotactile residual branch to a frozen backbone \citep{wang2026vitar}, while ResTacVLA applies residual adaptation in feature space \citep{zhang2026feeling}.
Tactile feedback can also guide action execution through inference-time adaptation.
FBFM introduces asynchronous feedback during flow-matching execution without additional training \citep{li2026fbfm}, while inference-time policy steering adjusts a fixed policy using vision and touch \citep{wu2026inference}.
Complementary approaches use touch as training supervision without requiring tactile input at deployment \citep{gubernatorov2026hapticvla,cheng2026tau}.
These approaches enable tactile feedback to influence action generation and execution. In \method{}, a cached, history-aware visual representation connects a slow visual policy to a fast tactile editor, which combines visual memory with recent touch to refine the ongoing action chunk. Predictive supervision encourages these representations to capture evolving interaction dynamics, while auxiliary predictors are used only during training. Table~\ref{tab:related-work-comparison} compares T-WLA with closely related methods and demonstrates their difference.

%% file: section/03_methodology.tex
\section{Methodology}
\label{sec:method}

\subsection{Overview}
\label{sec:method-overview}

During contact-rich manipulation, high-frequency changes in contact pressure or location may require the robot to adjust its motion before the next visual update.
Tactile feedback captures these local changes, but deciding how to respond also depends on the task instruction and visual information about the ongoing interaction.
To retain this information between visual updates, \method{} shares a history-aware visual memory between a visual policy and a lightweight tactile editor.
\begin{figure*}[!t]
\centering
\includegraphics[width=.92\linewidth]{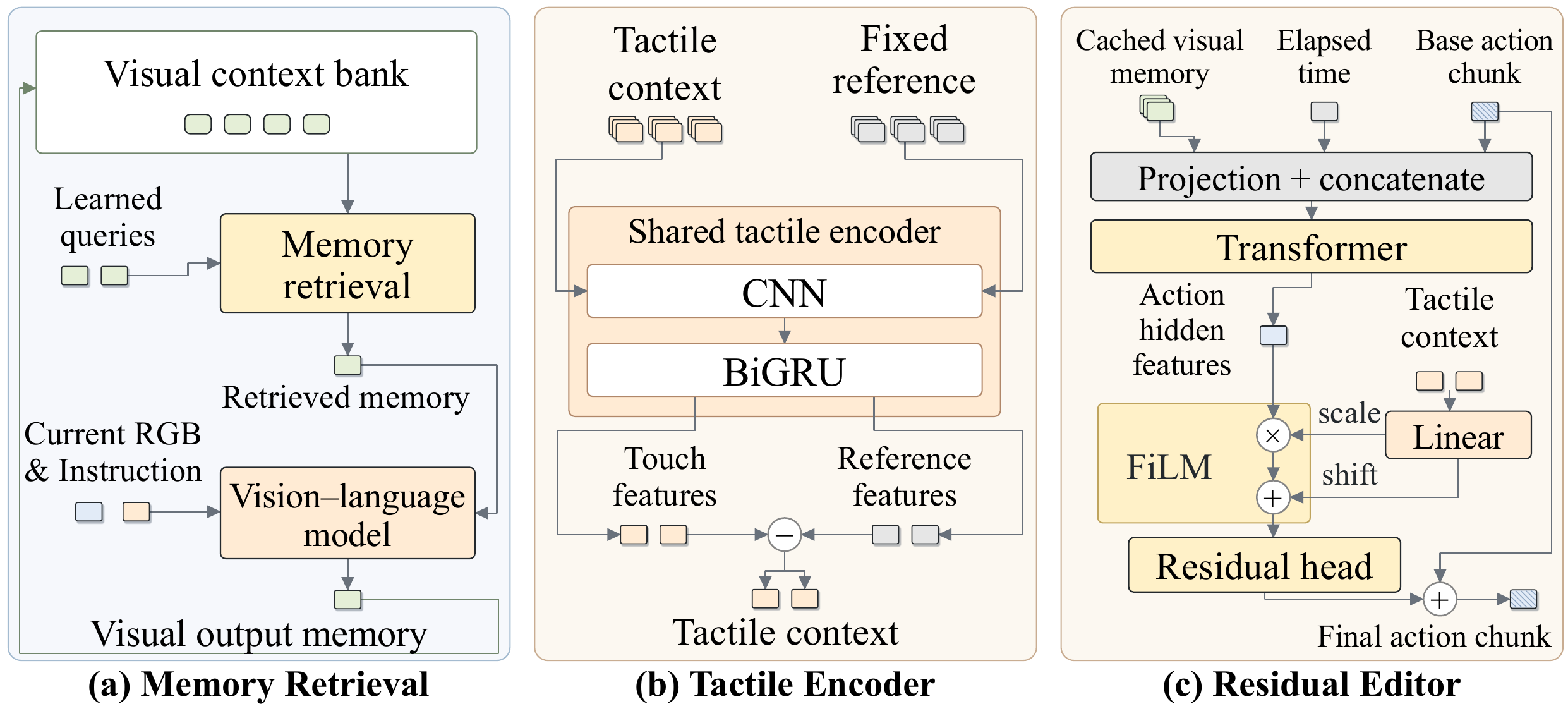}
\caption{
\textbf{The memory-to-editor interface.}
(a) Retrieved history is combined with the current observation and instruction.
(b) Observed tactile history is encoded relative to a fixed reference.
(c) Touch modulates action features conditioned on the cached visual memory, base chunk, and elapsed time.
Module details are in Appendices~\ref{app:visual-memory}--\ref{app:contact-editor}.
}
\label{fig:architecture-2}
\end{figure*}
The editor combines this cached memory with recent tactile observations to asynchronously correct the remaining actions planned by the visual policy, instead of rerunning the visual branch.

As shown in Fig.~\ref{fig:architecture}, \method{} processes RGB observations $I_t$ and higher-frequency tactile observations $T_{t,\tau}$, where $t$ indexes visual updates and $\tau$ indexes tactile updates between them.
At each visual update, the visual branch produces memory $M_\V^t$ and an $H$-step base action chunk $A_t^{\base}=f_{\act}(M_\V^t)$.
Both are cached while the tactile editor updates the remaining actions:

\par\noindent
\begin{minipage}{\columnwidth}
\centering
\renewcommand{\eqnannotationfont}{\sffamily\footnotesize}
\vspace*{1.4em}

\begin{equation}
\begin{aligned}
\widehat A_{t,\tau}
&=\eqnmarkbox[NavyBlue]{twla-act}{f_{\act}}
  \!\bigl(\eqnmarkbox[OliveGreen]{twla-memory-v}{M_\V^t}\bigr)
\\[2.4em]
&\quad+\eqnmarkbox[BurntOrange]{twla-editor}{\mathcal E}
  \!\bigl(
    \eqnmarkbox[OliveGreen]{twla-memory-t}{M_\V^t},
    \eqnmarkbox[NavyBlue]{twla-base}{A_t^{\base}},
    \eqnmarkbox[Plum]{twla-touch}{Z_{t,\tau}},
    \eqnmarkbox[WildStrawberry]{twla-time}{\Delta t_{t,\tau}}
  \bigr).
\end{aligned}
\label{eq:control-overview}
\end{equation}

\annotate[xshift=1.9em,yshift=0.7em]{above,left}
  {twla-act}{Base-action generation}
\annotate[yshift=0.7em]{above,right}
  {twla-memory-v}{Shared visual memory}
\annotate[yshift=0.7em]{above,left}
  {twla-base}{Cached base action chunk}
\annotate[yshift=0.7em]{above,right}
  {twla-touch}{Tactile context}
\annotate[yshift=-0.7em]{below,left}
  {twla-editor}{Tactile editor}
\annotate[yshift=-0.7em]{below,left}
  {twla-time}{Elapsed time}

\par\vspace*{1.2em}
\end{minipage}
\par

Here, $Z_{t,\tau}$ summarizes recent tactile history, $\Delta t_{t,\tau}$ is the elapsed time since the current active chunk started, and $\mathcal E$ returns an effective residual $\Delta A_{t,\tau}$ for unexecuted actions.

\subsection{History-Aware Visual Memory}
\label{sec:visual-memory}

A current image may obscure information that was visible earlier or leave the recent approach to contact ambiguous~\citep{wang2025vggt, zhou2026page, zhou2023dynpoint}.
We retain a compact visual history so that both planning and correction can use earlier observations without repeatedly encoding the full image sequence.
As illustrated in Fig.~\ref{fig:architecture-2}(a), learned queries $Q_\V$ retrieve input memory from a bank $\mathcal B_\V^{t-1}$ containing the last $K$ sets of visual output-memory tokens:
\begin{equation}
M_{\V,\inp}^{t}
=
\mathcal M_\V\!\left(\mathcal B_\V^{t-1},Q_\V\right).
\label{eq:memory-retrieval}
\end{equation}
The visual backbone then combines this history with current image tokens $E_t$, instruction tokens $L_\ell$, field-of-view tokens $c_t$, and state tokens $S_t$:
\begin{equation}
M_\V^t
=
f_\V\!\left(E_t,L_\ell,c_t,S_t,M_{\V,\inp}^{t}\right).
\label{eq:visual-memory}
\end{equation}
The updated visual memory $M_\V^t$ conditions the action expert $f_{\act}$ in Eq.~\ref{eq:control-overview}.
The bank update $\mathcal B_\V^t=\operatorname{Append}_K(\mathcal B_\V^{t-1},M_\V^t)$ appends the new memory while retaining at most the $K$ most recent memory sets.
The tactile editor uses the updated visual memory $M_\V^t$, which combines visual history with the current observation and task instruction. The visual branch is trained to predict demonstrated action chunks using conditional flow matching~\citep{lipman2022flow}, with future-image prediction providing supervision in Section~\ref{sec:predictive-memory}. Details are specified in Appendices~\ref{app:visual-memory} and ~\ref{app:losses}.

\begin{figure*}[!t]
  \centering
  \includegraphics[width=.92\linewidth]{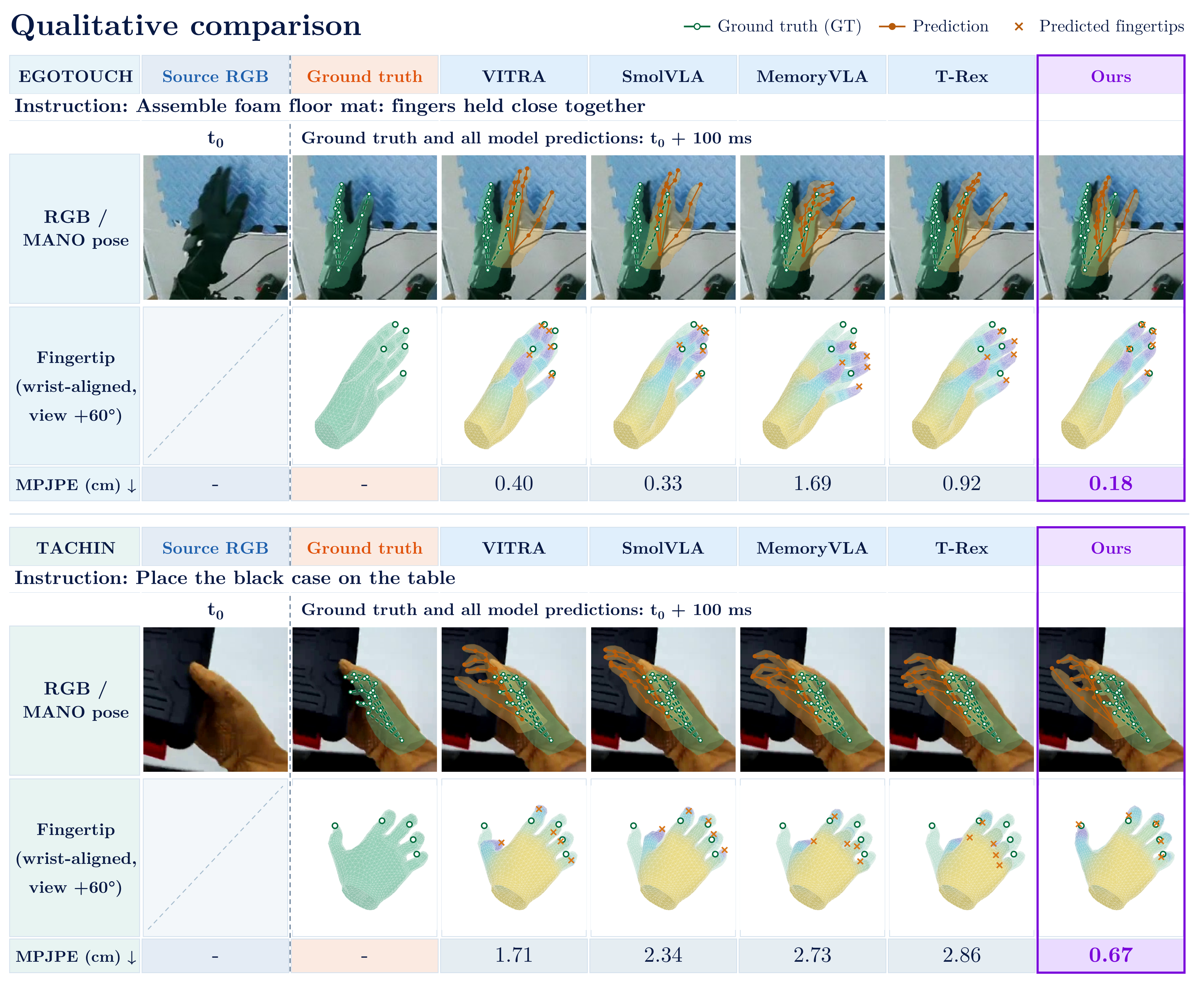}
  \vspace{-0.5cm}
 \caption{
\textbf{Qualitative comparison on EgoTouch and Tachin.}
  We show the tactile-augmented variants of VITRA, SmolVLA, and MemoryVLA, along with T-Rex and T-WLA using their native tactile interfaces, as evaluated in Table~\ref{tab:tactile-prediction}. 
  For each dataset, we show the source RGB frame at $t_0$, ground truth, and predictions from each of the models at $t_0 + 100\,\mathrm{ms}$.
RGB overlays depict ground-truth hand poses in green and predictions in orange.
Fingertip-detail views show wrist-aligned hands rendered from a $+60^\circ$ visualization view, with ground-truth hand mesh in green and predicted fingertips in orange crosses.
MPJPE is reported in centimeters ($\downarrow$: lower is better).
\method{} obtains the lowest MPJPE on both displayed examples.}
  \label{fig:qualitative_main}
\end{figure*}

\subsection{Asynchronous Tactile Correction}
\label{sec:tactile-interface}

The editor asynchronously refines an active plan through residual corrections rather than generating a new action plan from touch alone.
Visual memory supplies retained task and scene information; the base action chunk specifies what is being executed; and elapsed time identifies the remaining action positions as in Eq.~\ref{eq:control-overview}.
Between visual updates, $M_\V^t$ and $A_t^{\base}$ remain fixed, while each tactile update $\tau$ runs only the lightweight tactile encoder and residual editor, enabling efficient corrections without rerunning the visual branch.
This formulation assumes that the base plan remains a useful reference between visual updates.
Here, asynchronous correction denotes edits between visual replans; it does not require a new visual forward pass for each edit. In the measured command pipeline, only actions not yet dispatched are replaceable, and a new visual update supplies a new memory and base chunk. 

\textbf{Tactile context.}
We encode recent causally available tactile frames $\mathcal H_{t,\tau}$ as $Z_{t,\tau}=\operatorname{Enc}_{\T}(\mathcal H_{t,\tau})$.
A temporal window exposes contact evolution that a single pressure map cannot resolve.
Figure~\ref{fig:architecture-2}(b) shows the shared encoder and reference subtraction; Appendix~\ref{app:tactile-encoding} provides details.

\input{table/tab-1}

\textbf{Plan-conditioned correction.}
As illustrated in Fig.~\ref{fig:architecture-2}(c), a small Transformer combines the cached visual memory, base action chunk, and elapsed time to produce action features at each tactile update.
FiLM~\citep{perez2018film} uses the tactile context to scale and shift these features, which a residual head converts into action corrections.
After masking, the corrections $\Delta A_{t,\tau}$ affect only valid remaining actions; a hand with missing tactile input receives zero correction. 
In deployment, the editable set excludes already-dispatched commands.
The residual is added to the original cached base action chunk, yielding $\widehat A_{t,\tau}=A_t^{\base}+\Delta A_{t,\tau}$ as in Eq.~\ref{eq:control-overview}, rather than accumulated on previously corrected actions.
Appendix~\ref{app:contact-editor} describes the editor architecture in detail.
With the visual branch frozen, we supervise the corrected actions using a group-balanced Smooth L1 loss against demonstrations and regularize the predicted residuals with a magnitude penalty.
These objectives train the tactile encoder and residual editor to correct action errors while discouraging unnecessarily large corrections. Definitions of objectives are provided in Appendix~\ref{app:losses}.

\subsection{Predictive Supervision and Training}
\label{sec:predictive-memory}
\label{sec:training-strategy}

Action supervision specifies the demonstrated command, but may not distinguish states with different subsequent visual or contact outcomes.
We add future-image and future-pressure prediction to supervise information about these outcomes in the representations used for control.
Both prediction heads are training-only auxiliaries, not inference-time planners.

\textbf{Complementary prediction targets.}
The image predictor conditions on current image tokens and output visual memory to predict future latents supplied by a frozen image encoder.
During editor training, tactile context also conditions this predictor through a learned adapter.
The adapter changes only the auxiliary prediction path, leaving the cached memory used for control unchanged.
Future-pressure prediction instead supervises subsequent contact, conditioned on visual memory, tactile context, and upcoming edited actions.
The action condition is detached: pressure-prediction gradients update tactile features through their conditioning path, but do not update the residual readout through the edited actions.
Predictor conditions, gradient routes, and normalization are specified in Appendix~\ref{app:predictive}; future targets never enter the controller.

\textbf{Staged optimization.}
The visual action expert and predictive heads use conditional flow matching~\citep{lipman2022flow}.
We first train the visual branch with action and image-prediction supervision, and separately pretrain the tactile encoder through contrastive learning on synchronized tactile and hand-pose histories from EgoTouch.
We then freeze the visual branch and use its cached visual memory and base action chunks to train the tactile encoder and residual editor with a group-balanced Smooth L1 action loss, a residual-magnitude penalty, and the two predictive objectives.
The visual predictor remains frozen, while its tactile conditioning adapter and the pressure predictor are trained.
Objective definitions and custom reductions are provided in Appendices~\ref{app:losses} and~\ref{app:predictive}.
Stage-specific training details are given in Appendix~\ref{app:training-stages}; joint fine-tuning is not part of the reported additional experiments.
Both prediction heads are removed at inference.

%% file: table/tab-1.tex
\begin{table*}[!t]
\centering
\caption{Tactile-conditioned future hand-motion prediction.
Home--Outdoor denote the five EgoTouch environments, and Avg.\ reports their macro-average.
Overall reports results on the Tachin test split.
Metrics are evaluated at a common prediction horizon and averaged within each sequence before aggregation.
Lower is better.
Gray shading highlights \method{}.
\textbf{Bold} and \underline{underlining} indicate the best and second-best distinct errors within each metric and row, including ties.
Tactile Freq.\ denotes the reported tactile-correction rate in Hz.
}
\label{tab:tactile-prediction}
\footnotesize
\setlength{\tabcolsep}{2pt}
\renewcommand{\arraystretch}{1.12}
\resizebox{\linewidth}{!}{%
\begin{tabular}{@{}l*{3}{cccc>{\columncolor[gray]{0.93}}c}@{}}
\toprule
& \multicolumn{5}{c}{Wrist translation error (cm, $\downarrow$)}
& \multicolumn{5}{c}{Wrist rotation geodesic error ($^\circ$, $\downarrow$)}
& \multicolumn{5}{c}{Wrist-relative finger-joint MPJPE (cm, $\downarrow$)} \\
\cmidrule(lr){2-6}\cmidrule(lr){7-11}\cmidrule(lr){12-16}
Evaluation split
& \shortstack{VITRA\\+ tactile} & \shortstack{SmolVLA\\+ tactile}
& \shortstack{MemoryVLA\\+ tactile} & T-Rex & \method{}
& \shortstack{VITRA\\+ tactile} & \shortstack{SmolVLA\\+ tactile}
& \shortstack{MemoryVLA\\+ tactile} & T-Rex & \method{}
& \shortstack{VITRA\\+ tactile} & \shortstack{SmolVLA\\+ tactile}
& \shortstack{MemoryVLA\\+ tactile} & T-Rex & \method{} \\
\midrule
Home
& \underline{4.54} & 4.55 & 4.65 & 4.63 & \textbf{4.36}
& \underline{6.29} & \underline{6.29} & 6.34 & \underline{6.29} & \textbf{6.18}
& \underline{0.51} & \underline{0.51} & 1.37 & 1.17 & \textbf{0.36} \\
Workb.
& \underline{3.71} & 3.74 & 3.78 & 3.76 & \textbf{3.63}
& \underline{6.22} & \underline{6.22} & 6.39 & 6.29 & \textbf{5.83}
& 0.51 & \underline{0.50} & 1.37 & 1.13 & \textbf{0.33} \\
Office
& \underline{2.84} & 2.88 & 2.96 & 2.95 & \textbf{2.73}
& \underline{4.81} & 4.87 & 4.88 & 4.84 & \textbf{4.79}
& \underline{0.45} & 0.46 & 1.17 & 1.07 & \textbf{0.31} \\
Retail
& \underline{3.98} & \underline{3.98} & 4.08 & 4.05 & \textbf{3.82}
& 6.86 & \underline{6.85} & 6.89 & 6.87 & \textbf{6.66}
& 0.51 & \underline{0.50} & 1.34 & 1.17 & \textbf{0.36} \\
Outdoor
& 5.34 & \underline{5.33} & 5.41 & 5.41 & \textbf{5.24}
& \underline{7.15} & 7.16 & 7.20 & 7.17 & \textbf{7.03}
& \underline{0.53} & \underline{0.53} & 1.46 & 1.18 & \textbf{0.40} \\
Avg.
& \underline{4.08} & 4.10 & 4.18 & 4.16 & \textbf{3.96}
& \underline{6.27} & 6.28 & 6.34 & 6.29 & \textbf{6.10}
& \underline{0.50} & \underline{0.50} & 1.34 & 1.14 & \textbf{0.35} \\
\midrule
Overall
& \underline{1.48} & 1.50 & 1.55 & 1.54 & \textbf{1.39}
& 4.55 & \underline{4.20} & 4.39 & 4.25 & \textbf{4.17}
& \underline{0.63} & 1.28 & 1.87 & 1.08 & \textbf{0.46} \\
\midrule
Tactile Freq.
& 58.2 & 69.0 & 15.0 & 17.16 & 119.1
& 58.2 & 69.0 & 15.0 & 17.16 & 119.1
& 58.2 & 69.0 & 15.0 & 17.16 & 119.1 \\
\bottomrule
\end{tabular}%
}
\end{table*}

%% file: section/06_evaluation.tex
\section{Experiments}
\label{sec:evaluation}

We organize our experiments around four questions. First, does combining visual memory with high-rate tactile feedback improve future hand-motion prediction on human-interaction datasets (Section~\ref{sec:tactile-prediction})? Second, does the tactile branch improve closed-loop success on real contact-rich robot tasks under a matched visual backbone and training budget (Section~\ref{sec:robot-evaluation})? Third, which components---tactile input, recurrent memory, memory capacity, and predictive supervision---account for the improvement (Section~\ref{sec:ablation})? Finally, appendices~\ref{app:visual-only} and~\ref{app:tactile-input} report visual-only action prediction results and ablations of tactile input and update timing, respectively.

\begin{figure*}[!t]
  \centering
  \includegraphics[width=\linewidth]{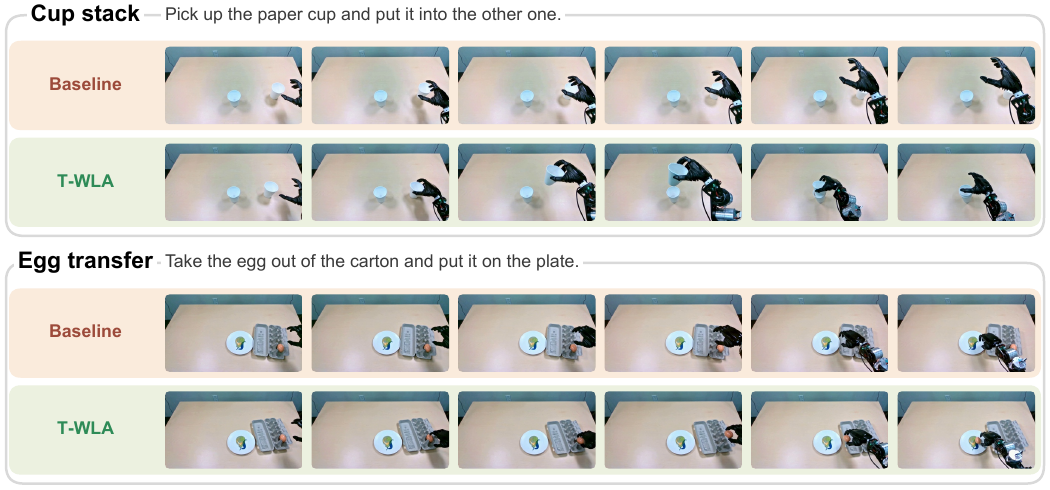}
  \caption{Representative real-robot trials for cup stack and egg transfer. For each task, the upper row shows the baseline and the lower row shows T-WLA with tactile editing. Frames are ordered from left to right.}
  \label{fig:robot}
\end{figure*}

\subsection{Experimental Setup}
\label{sec:experimental-setup}

\textbf{Datasets and tasks.}
For tactile-conditioned future hand-motion prediction, we evaluate on EgoTouch from TouchAnything~\citep{zhou2026touchanything} and the Tachin glove dataset~\citep{tachintech}. EgoTouch is evaluated in its five native environments---Home, Workbench, Office, Retail, and Outdoor---whereas Tachin is reported over its complete held-out test set. For visual-only prediction, we use GigaHands~\citep{fu2025gigahands}, OakInk2~\citep{,zhan2024oakink2}, and TACO~\citep{cheng2026taco}. GigaHands is reported over its five official activity families, while OakInk2 and TACO are each evaluated over their complete held-out test sets. Our real-robot benchmark contains four contact-rich tasks: egg transfer, single-sheet pickup, paper-cup nesting, and paper-bag dragging. Appendix~\ref{app:robot-setup} describes the robot platform, teleoperation setup, and demonstration collection.

\textbf{Baselines and controlled comparison.}
For tactile-conditioned prediction, we compare with VITRA~\citep{li2025scalable}, SmolVLA~\citep{shukor2025smolvla}, MemoryVLA~\citep{shi2026memoryvla}, and T-Rex~\citep{niu2026t}. When a baseline does not natively accept touch, its ``+ tactile'' variant receives the same synchronized tactile stream and uses the same task-specific tactile encoder, prediction head, and optimization budget as the other adapted baselines. T-Rex and \method{} retain their native multi-rate interfaces. Appendix~\ref{app:protocol} specifies the action representation, temporal alignment, metric definitions, evaluation masks, and aggregation procedure. 

\subsection{Tactile-Conditioned Action Prediction}
\label{sec:tactile-prediction}

\textbf{Tactile-conditioned hand-motion prediction.}
Table~\ref{tab:tactile-prediction} shows that \method{} achieves the lowest reported errors across all three metrics in every EgoTouch environment and on Tachin.
The clearest advantage, however, is in finger articulation: wrist-relative MPJPE decreases by approximately $30.0\%$ on EgoTouch and $27.0\%$ on Tachin relative to the strongest reported baseline for each dataset.
Because this metric measures joint positions relative to the wrist, the improvement is not simply a consequence of predicting the global wrist position more accurately.
Wrist translation and rotation also improve, but more modestly.
This contrast is consistent with the intended role of the tactile editor: refining fine-grained hand motion within an ongoing visual plan, rather than replacing the plan altogether.
The same pattern across both datasets, together with lower errors in all five EgoTouch environments, indicates that the benefit is not confined to a single environment.
The accuracy improvements also coexist with a higher effective tactile-correction rate.
We benchmark the effective tactile-correction rate of each method on a single NVIDIA H100 GPU.
Under our multi-rate protocol, \method{} reaches $119.1$\,Hz, compared with $69.0$\,Hz for the fastest reported baseline.
Thus, the lightweight correction pathway combines more frequent tactile updates with lower prediction errors, rather than requiring a lower update rate to obtain better predictions.
This accuracy--update-rate combination is particularly relevant to asynchronous visuotactile control, where contact-driven corrections are needed between slower visual updates.
Appendix~\ref{app:appendix_qualitative} provides additional visualizations on EgoTouch and Tachin.

\input{table/tab-2}

\input{table/tab-3}

\subsection{Real-Robot Evaluation}
\label{sec:robot-evaluation}
Next, we test whether the prediction improvements translate into closed-loop manipulation on four tasks: egg transfer, single-sheet pickup, paper-cup nesting, and paper-bag dragging.
These tasks are contact-sensitive: egg transfer involves small in-hand slip that is difficult to observe visually; sheet pickup and cup nesting require a secure grip without crumpling or crushing the object; and bag dragging requires maintaining a grasp on the loaded bag's rim under changing fingertip loads.
Together, they test whether tactile feedback can refine visually guided actions using contact information that images alone may not resolve.
As shown in Table~\ref{tab:robot-success}, \method{} with tactile input achieves a mean success rate of $51.0\%$ with only 30 teleoperated demonstrations per task.
Before robot adaptation, \method{}, including its tactile editor, is trained only on human-hand data.
Even without tactile input, \method{} reaches $44.25\%$ mean success, compared with $16.75\%$ for SmolVLA, which shares components with our visual branch.
VITRA and MemoryVLA achieve no successful trials under the same demonstration budget.
These closed-loop results complement the finger-motion prediction improvements in Table~\ref{tab:tactile-prediction}.
Relative to \method{} without tactile input, tactile feedback improves success by 9 percentage points each on single-sheet pickup and paper-bag dragging, 5 points on egg transfer, and 4 points on paper-cup nesting.
A thin sheet requires a precise pinch with enough force to prevent slipping but not enough to crumple it, so this improvement is consistent with tactile feedback helping regulate a delicate grasp.
The gains in egg transfer and paper-bag dragging may reflect the value of sensing changes in grasp stability and fingertip load.
Paper-cup nesting improves from 16\% to 20\%. The task requires a secure grasp that avoids crushing the cup, followed by accurate alignment and placement.
Grip-related corrections may be insufficient if failures occur during these other stages, although the aggregate success rate does not identify the limiting stage.
In observed SmolVLA failures, the hand often closes away from the target, leaving little contact feedback to guide a correction.
This may explain its smaller tactile gain when the base action fails to establish contact.
Appendix~\ref{app:realrobot-qualitative} shows additional examples of paper-bag dragging and single-sheet pickup.
Appendix~\ref{app:hardware-timing} reports visual replanning rates, tactile editing rates, and host-side latency from a separate hardware-in-the-loop profile of the command pipeline.


\subsection{Ablation Studies}
\label{sec:ablation}

\textbf{Ablation design.}
Table~\ref{tab:ablation} evaluates the roles of tactile feedback, visual history, residual prediction, and predictive supervision.
\emph{Without tactile input} withholds tactile observations while retaining visual history.
\emph{Without recurrent memory} disables history retrieval in the visual backbone while retaining tactile editing.
\emph{Without tactile and recurrent memory} disables history retrieval and evaluates the resulting base actions without tactile editing.
\emph{Shorter memory} reduces the visual history bank from eight to four stored updates.
\emph{Conditional diffusion residual head} replaces direct residual regression with a conditional diffusion model using five DDIM steps.
\emph{Without world prediction (T), (V), and (T+V)} remove future-pressure, future-image, or both auxiliary objectives during editor training, respectively, while retaining the pretrained visual backbone.

\textbf{Effects of tactile feedback and visual history.}
Removing tactile input has its largest relative effect on finger articulation, increasing finger-joint MPJPE by $22.9\%$ on EgoTouch and $8.7\%$ on Tachin. 
Removing recurrent visual history particularly affects wrist motion on Tachin, where translation and rotation errors rise by $21.6\%$ and $9.4\%$, respectively. 
Halving the history capacity raises relative error by $2.3\%$ on average across the six reported metrics.
In the matched $2\times2$ control, visual history lowers prediction errors both before and after editing, and tactile editing lowers errors with either history setting. 
Compared with disabling both tactile input and recurrent visual memory, enabling both reduces finger-joint MPJPE by 40.6\% on EgoTouch and 18.6\% on Tachin.

\textbf{Residual prediction and auxiliary supervision.}
Under matched conditioning and training budgets, direct regression yields lower errors than the diffusion head on all three metrics across both datasets, reducing finger-joint MPJPE by 26.9\% on EgoTouch and 25.4\% on Tachin.
Removing either predictive objective raises all six reported errors. 
Removing both raises relative error by $1.0\%$ on average, indicating a modest contribution from predicting future images and pressure during editor training.

%% file: table/tab-2.tex
\begin{table}[!t]
\centering
\caption{Closed-loop success rates (\%, $\uparrow$) on four contact-rich
robot tasks after adaptation with 30 teleoperated demonstrations per task.
Avg. denotes the unweighted mean across tasks.
\textbf{Bold}/\underline{underline} indicate the best/second-best
distinct results in each column, including ties.
Gray highlights \method{}.}
\label{tab:robot-success}
\small
\resizebox{\linewidth}{!}{%
\begin{tabular}{@{}lrrrrr@{}}
\toprule
Controller
& \shortstack{Egg\\Transfer}
& \shortstack{Single-Sheet\\Pickup}
& \shortstack{Paper-Cup\\Nesting}
& \shortstack{Paper-Bag\\Dragging}
& Avg. \\
\midrule
SmolVLA
& 0 & 28 & 0 & 39 & 16.75 \\
SmolVLA + Tactile
& 0 & 30 & 1 & 37 & 17.0 \\
T-Rex
& 4 & 3 & \textbf{22} & 63 & 23.0 \\
\method{} w/o tactile
& \underline{17} & \underline{69} & 16
& \underline{75} & \underline{44.25} \\
\rowcolor{gray!15}
\method{}
& \textbf{22} & \textbf{78} & \underline{20}
& \textbf{84} & \textbf{51.0} \\
\bottomrule
\end{tabular}%
}
\end{table}

%% file: table/tab-3.tex
\begin{table*}[!t]
  \par
  \caption{Ablation of tactile input, memory, and predictive
  supervision. World-prediction ablations remove auxiliary objectives
  during editor training, not backbone pretraining. Lower is better
  for every metric. \textbf{Bold}/\underline{underline} indicate the
  best/second-best distinct results.}
  \label{tab:ablation}
  \centering
  \small
  \setlength{\tabcolsep}{5pt}
  \begin{tabular}{@{}lrrr rrr@{}}
  \toprule
  & \multicolumn{3}{c}{EgoTouch}
  & \multicolumn{3}{c}{Tachin} \\
  \cmidrule(lr){2-4}\cmidrule(l){5-7}
  Variant
  & Trans. (cm) & Rot. ($^\circ$) & Joint (cm)
  & Trans. (cm) & Rot. ($^\circ$) & Joint (cm) \\
  \midrule
  Without tactile \& memory
      & 6.010 & 8.835 & 0.589
      & 1.734 & 4.929 & 0.565 \\
  Without tactile input
      & 4.010 & 6.200 & 0.430
      & 1.450 & 4.400 & 0.500 \\
  Conditional diffusion residual head
      & 4.484 & 6.868 & 0.479
      & 1.456 & 4.419 & 0.617 \\
  \midrule
  Without recurrent memory
      & 4.070 & 6.280 & 0.370
      & 1.690 & 4.560 & 0.490 \\
  Shorter memory ($K_{\mathrm{short}}<K$)
      & 3.980 & 6.150 & 0.360
      & 1.480 & 4.310 & \textbf{0.460} \\
  \midrule
  Without world prediction (T+V)
      & 3.963 & 6.113 & 0.353
      & 1.413 & 4.283 & 0.463 \\
  Without world prediction (T)
      & \underline{3.962}
      & \underline{6.112}
      & \underline{0.352}
      & \underline{1.412}
      & \underline{4.263}
      & \underline{0.462} \\
  Without world prediction (V)
      & 3.963 & 6.113 & 0.353
      & 1.423 & 4.273 & 0.463 \\
  \midrule

  \rowcolor[gray]{0.93}
  \textbf{\method{}}
      & \textbf{3.960} & \textbf{6.100} & \textbf{0.350}
      & \textbf{1.390} & \textbf{4.170} & \textbf{0.460} \\
  \bottomrule
  \end{tabular}
  \par
  \vspace{-0.6cm}
\end{table*}

%% file: section/07_conclusion.tex
\section{Conclusion}
We presented \method{}, a multi-rate visuotactile architecture that caches history-aware visual memory to guide high-frequency tactile corrections.
A slow visual policy generates base action chunks, while a lightweight tactile editor reuses the cached visual memory to refine ongoing actions at a higher frequency.
Future-image and future-pressure prediction regularize the control representations during training, complementing action supervision with interaction dynamics.
Extensive results show that joint visual-tactile modeling and predictive supervision improve contact-rich manipulation.

%% file: section/ai_disclosure.tex
\section*{LLM Usage Statement}

We used ChatGPT (OpenAI) to polish the wording of the manuscript and Codex
(OpenAI) to format the LaTeX files for CVPR. We did not use LLMs to write
or modify research code. The authors take responsibility for the final
manuscript.

%% file: section/09_appendix.tex
\section{Action Foundation Model}
\label{app:action-foundation}

General-purpose manipulation policies increasingly adapt large pretrained backbones to action prediction \citep{wu2026pragmatic,ye2026dreamzero, yuan2026fastwam,black2024pi_0,pertsch2025fast,kim2025fine,team2024octo,liu2025rdt,bjorck2025gr00t,team2025gemini,shukor2025smolvla,o2024open,driess2023palm,brohan2022rt,reed2022generalist,cheang2024gr,zhu2023vima,bousmalis2023robocat,li2024vision,wang2024scaling}.
Vision--language--action (VLA) models follow this approach by predicting robot commands from visual observations and language instructions.
RT-2 \citep{brohan2023rt} represents actions as tokens to jointly train on robot trajectories and web data, while OpenVLA \citep{kim2024openvla} fine-tunes a pretrained vision--language backbone on a large collection of real robot demonstrations.
More recent work expands the data and supervision available for policy learning.
$\pi_{0.5}$ \citep{intelligence2025pi_} combines robot trajectories, web data, and subtask supervision to support generalization to unseen environments.
SmolVLA \citep{shukor2025smolvla} emphasizes efficient training with community demonstrations, while VITRA \citep{li2025scalable} learns a hand-motion prior from human activity videos.
World--action models (WAMs) \citep{hou2026worldmodelrobotlearning} further incorporate future-state prediction to learn representations of how the environment evolves.
FLARE \citep{team2025flare} predicts future observation embeddings rather than reconstructing future pixels.
DreamZero \cite{ye2026dreamzero} jointly predicts future video and robot actions.
Fast-WAM \citep{yuan2026fastwam} uses predictive learning during training while removing the predictor at deployment.
These methods motivate the use of future prediction to learn control representations. 
In contact-rich manipulation, local contact conditions can change before the next visual replan. 
At each visual update, \method{} caches a history-aware visual representation that the tactile editor combines with incoming touch to predict residual corrections to the active action chunk.

\input{table/tab-3-1}

\section{Module Implementation}
\label{app:modules}

This appendix expands the implementation of the operators introduced in the main text. We retain $M_{\mathrm V}^t$ for the final visual memory, $Z_{t,\tau}$ for the reference-centered tactile context, and $\Delta A_{t,\tau}$ for the effective action correction. Numerical specifications below refer to the reference implementation; the contact-filtered variant is identified where it changes an operation or its valid supervision set.

\subsection{Visual Memory Retrieval}
\label{app:visual-memory}
\textbf{Queries, reference slots, and historical tokens.}
Let $t$ denote the current visual update. The retrieval module has $S=8$ learned query tokens $Q_{\mathrm V}=[Q_{\mathrm V,1},\ldots,Q_{\mathrm V,S}] \in\mathbb R^{S\times d}$, each of width $d=960$. Its history bank stores up to $K=8$ previous visual output-memory sets $M_{\mathrm V}^{t-k}\in\mathbb R^{S\times d}$, where $k=1$ denotes the most recent set. For slot $j \in \{1,\ldots,S\}$, $e_j^{\mathrm{slot}}\in\mathbb R^d$ identifies the slot; for age $k \in \{1,\ldots,K\}$, $e_k^{\mathrm{age}}\in\mathbb R^d$ identifies how many visual updates ago the set was produced.
The query stream starts from $Q_{\mathrm V}$, while the key/value stream contains learned reference slots and available historical tokens:
\begin{align}
 r_j &= Q_{\mathrm V,j}+e_j^{\mathrm{slot}}, \\
 b_{k,j}^{t} &= M_{\mathrm V,j}^{t-k}
                 +e_j^{\mathrm{slot}}+e_k^{\mathrm{age}},
 \qquad k\in\{1,\ldots,K\}.
 \label{eq:app-retrieval-tokens}
\end{align}
Thus, $r_j$ is the reference token for slot $j$, reusing the query parameter rather than introducing a separate learned vector;
$b_{k,j}^{t}$ is the token from slot $j$ of the output memory written $k$ visual updates earlier. Queries attend to all reference tokens and valid historical tokens; unavailable history is masked. Slot and age embeddings are added only during retrieval and are not stored in the bank. Age is measured in visual updates, not tactile events or action-flow steps.

\textbf{Masking and output extraction.}
Unused bank positions are zeroed and excluded by the attention mask; reference slots remain valid even when the bank is empty. Two retrieval blocks use pre-normalized cross-attention and SwiGLU feed-forward layers, with 15 attention heads and an intermediate width of 2560. The blocks use Q/K RMS normalization, a learned attention-logit scale, and zero-initialized
$\tanh$ gates on their residual branches. 
The retrieved input memory is placed after the image, field-of-view, language, and state tokens in the visual backbone input. The final $M_{\mathrm V}^t$ is read directly from the last decoder layer at these eight memory positions, without an additional readout projection. The bank stores this observation-contextualized output, not the retrieved input memory. Episode boundaries reset the bank; a visual event writes one new memory set, independently of the number of action-flow integration steps.

\subsection{Tactile Encoding}
\label{app:tactile-encoding}

\textbf{Sensor processing and temporal encoding.}
In the EgoTouch reference implementation, each hand is represented by its eight most recent causally available $21\times21$ pressure maps. Invalid temporal observations are zeroed, and the left-hand maps are flipped along the spatial horizontal axis to establish a common hand convention. Pressure and the spatial sensor-validity mask form two input channels. A shared CNN uses three $5\times5$ convolutional layers with 32 channels, padding 2, ReLU activations, and $2\times2$ max-pooling. The resulting 128-dimensional frame features enter a two-layer bidirectional GRU with hidden width 120 per direction. The final forward and initial backward states are concatenated and projected from 240 to 64 dimensions, followed by $\ell_2$ normalization. Both GRU directions operate within the observed history window; neither accesses a future tactile frame.

\textbf{Reference subtraction.}
Let $F_{\mathrm T}$ denote the shared tactile encoder, including its output normalization but excluding reference subtraction. At tactile update $(t,\tau)$, $\mathcal H_{t,\tau}^{r}$ is the ordered tactile history for hand $r\in\{\mathrm L,\mathrm R\}$, and $\mathcal H_{\mathrm{ref}}^{r}$ is its fixed, hand-specific reference history. Both are encoded with the same module $F_{\mathrm T}$. The reference-centered representation supplied to the tactile editor is
\begin{equation}
 Z_{t,\tau}^{r}
 =v_{t,\tau}^{r}\left[
 F_{\mathrm T}(\mathcal H_{t,\tau}^{r})
 -F_{\mathrm T}(\mathcal H_{\mathrm{ref}}^{r})\right].
 \label{eq:app-centered-touch}
\end{equation}
Here $v_{t,\tau}^{r}\in\{0,1\}$ equals one only when the encoded history passes its validity check and the most recent tactile observation is valid; otherwise, the representation is set to zero. In the contact-filtered variant, contact validity additionally affects the temporal mask used to encode the history, rather than introducing a separate gate after subtraction. Each encoder output is normalized before subtraction; the resulting difference is not normalized again.

The fixed reference is a raw pressure window, not a cached embedding. Actual and reference windows are concatenated along the batch dimension, encoded together on every forward pass, and separated before subtraction. Consequently, both encoding paths receive gradients when the encoder is trained. The reference-statistics routine pools valid training pressure observations from history windows and the associated future-pressure targets, and repeats the per-taxel mean across the eight reference frames.
These are fixed training-set statistics, not sample-specific future inputs to the controller. The exact statistics asset belongs to the training configuration.

\subsection{Residual Editing}
\label{app:contact-editor}

\textbf{Action features and tactile modulation.}
The editor maps the visual memory and base-action tokens to width 256. Visual tokens use RMS normalization followed by a bias-free projection; base-action tokens are standardized for their input projection. Modality, action-position, and elapsed-time embeddings distinguish the token roles. The elapsed-time projection receives $10\Delta t_{t,\tau}$, with elapsed time expressed in seconds. A two-layer pre-normalized Transformer uses eight heads, a feed-forward width of 1024, GELU activations, and zero dropout. Only its action-position outputs, denoted $H_{\mathrm{act}}$, enter the residual readout.

For each hand, a bias-free linear map from its 64-dimensional tactile context produces the FiLM scale and shift. Here $Z^r$ abbreviates the reference-centered context $Z_{t,\tau}^r$ defined above, and $H_{\mathrm{act}}$ denotes the action features associated with the active base chunk. The scale and shift have dimensions compatible with $H_{\mathrm{act}}$, allowing each hand's tactile context to modulate these action features before readout. With $U^r$ denoting an internal, unmasked correction, the readout is
\begin{align}
 U^r &= W_{\mathrm{out}}^r\!\left[
     \gamma^r(Z^r)\odot H_{\mathrm{act}}+\beta^r(Z^r)\right],\\
 \Delta A_{t,\tau} &= \sum_{r\in\{\mathrm L,\mathrm R\}}
                  m_{\mathrm{edit}}^r\odot U^r.
 \label{eq:app-effective-correction}
\end{align}
Here $\odot$ denotes element-wise multiplication. Each output projection $W_{\mathrm{out}}^r$ maps the modulated features to action coordinates; it
is bias-free and zero-initialized so that training starts with zero correction. The mask $m_{\mathrm{edit}}^r$ intersects the editable action positions, that hand's component mask, and hand validity. Thus $\Delta A_{t,\tau}$ retains its main-text meaning as the effective correction added to the cached base chunk, rather than the raw network output. Every update is relative to the same base chunk, not to a previously corrected chunk.

\textbf{Causal contact gate.}
The contact-filtered variant uses a hard gate computed from observed pressure history. Let $P_{j,r}$ be the pressure map of hand $r$ at history index $j$, let $\mathcal S_r$ contain its valid sensor sites, and let $\mathcal J_r$ contain its valid history indices. The gate is
\begin{equation}
 c^r=\mathbf 1\!\left[
  \max_{j\in\mathcal J_r}
  \frac{1}{5}\sum_{u\in\operatorname{Top5}(P_{j,r}|_{\mathcal S_r})}
           P_{j,r,u}\geq\eta
 \right],
 \qquad \eta=0.25.
 \label{eq:app-contact-gate}
\end{equation}
Here $\operatorname{Top5}$ returns the indices of the five largest valid pressure values. An empty valid history gives $c^r=0$. The reference threshold is expressed in the input pressure scale. The gate detects whether a valid history frame crosses the threshold; it does not require contact at the latest frame and is not learned from future contact labels.

The same $c^r$ is intersected with the hand's temporal tactile-validity mask, action-supervision mask, and future-pressure-validity mask. A gated-out hand therefore receives zero correction and retains its base action, but its excluded action entries are not counted as zero-error observations in the loss. The contact wrapper does not directly change the RGB target validity mask. Sample-level contact filtering and its evaluation denominator are described in Appendix~\ref{app:evaluation}.

\section{Learning Details}
\label{app:learning}

\subsection{Loss Function}
\label{app:losses}

\textbf{Visual action loss.}
The group-balanced operators for the visual expert and the tactile editor have different reduction orders. Let $b$, $j$, and $d$ index a minibatch
sample, an action position, and an action component. For visual velocity error $e_{b,j,d}$, validity mask $m^{\mathrm V}_{b,j,d}$, and component group $\mathcal D_g$, define
\begin{align}
 n_g &= \sum_{b,j}\sum_{d\in\mathcal D_g}m^{\mathrm V}_{b,j,d}, &
 \ell_g &= \frac{\sum_{b,j}\sum_{d\in\mathcal D_g}
                  m^{\mathrm V}_{b,j,d}e_{b,j,d}^{2}}{n_g}, \\
\mathcal L_{\mathrm{act}}^{\mathrm V}
 &=\frac{\sum_{g:n_g>0}w_g\ell_g}{\sum_{g:n_g>0}w_g}.
 \label{eq:app-visual-reduction}
\end{align}
Here $\ell_g$ is defined only for $n_g>0$; a minibatch with no valid groups contributes zero loss. The within-group mean is formed over the entire minibatch, not separately for each sample. Empty groups are excluded. The reference configuration uses equal group weights and no additional horizon weighting.

\textbf{Edited-action loss and residual penalty.}
Let $b$, $j$, and $d$ index a minibatch sample, an action position, and an action component, respectively, and let $\mathcal D_g$ be the components in group $g$. For corrected-action error $e_{b,j,d}=\widehat A_{b,j,d}-A^\star_{b,j,d}$, where $\widehat A$ is the corrected action chunk and $A^\star$ is the demonstration chunk, let $m^{\mathcal E}_{b,j,d}$ denote the effective valid, editable supervision mask and $n_{b,g}=\sum_j\sum_{d\in\mathcal D_g}m^{\mathcal E}_{b,j,d}$ the number of supervised entries in group $g$ for sample $b$. Write
$\mathcal G_b=\{g:n_{b,g}>0\}$ for the nonempty groups in sample $b$ and $\mathcal B_+=\{b:|\mathcal G_b|>0\}$ for the supervised samples. The minibatch estimator is
\begin{equation}
\mathcal L_{\mathrm{edit}}
 =\frac{1}{|\mathcal B_+|}\sum_{b\in\mathcal B_+}
  \frac{1}{|\mathcal G_b|}\sum_{g\in\mathcal G_b}
  \frac{\sum_j\sum_{d\in\mathcal D_g}
       m^{\mathcal E}_{b,j,d}\rho(e_{b,j,d})}{n_{b,g}},
 \label{eq:app-editor-reduction}
\end{equation}
where $\rho(e)=\tfrac12e^2$ for $|e|<1$ and $\rho(e)=|e|-\tfrac12$ otherwise. Thus the editor averages nonempty groups within each sample before averaging supervised samples. Invalid targets are replaced safely before elementwise evaluation so that invalid entries do not introduce non-finite losses.

In contrast, the residual penalty is a single mean over all effective
action entries:
\begin{equation}
 \widehat{\mathcal R}_{\Delta}
 =\frac{\sum_{b,j,d}m^{\mathcal E}_{b,j,d}
                     (\Delta A_{b,j,d})^2}
        {\sum_{b,j,d}m^{\mathcal E}_{b,j,d}},
 \label{eq:app-residual-reduction}
\end{equation}
where $\Delta A_{b,j,d}$ is the effective residual added to the base action at that entry. It does not rebalance groups or hands. When $\mathcal B_+$ is empty, or when the residual penalty has no valid entries, the corresponding loss is defined as zero. Contact filtering changes the valid sets in both expressions, rather than adding artificial zero-loss examples.

\begin{table}[!t]
\centering
\caption{Hand-motion action layout. Channel ranges are zero-based, half-open intervals. Joint channels store 21 three-dimensional local positions per hand.}
\vspace{-0.4cm}
\label{tab:app-action-layout}
\small
\setlength{\tabcolsep}{1pt}
\renewcommand{\arraystretch}{1.12}
\begin{tabular}{@{}lccc@{}}
\toprule
Component & Left & Right & Channels \\
\midrule
Wrist translation increment & $[0,3)$ & $[69,72)$ & 3 \\
Wrist Euler rotation increment & $[3,6)$ & $[72,75)$ & 3 \\
Palm-local joint coordinates & $[6,69)$ & $[75,138)$ & 63 \\
\midrule
Padding & \multicolumn{2}{c}{$[138,192)$} & --- \\
\bottomrule
\vspace{-0.8cm}
\end{tabular}
\end{table}

\subsection{Predictive Conditions}
\label{app:predictive}

\textbf{Tactile-conditioned visual prediction.}
During tactile editor training, a learned adapter maps the tactile context $Z_{t,\tau}$ to an additive update of the eight visual-memory tokens used by the frozen visual predictor. This update affects only the prediction condition; the cached visual memory used for action generation remains unchanged. Let $y$ denote the clean future-image latent, $\varepsilon$ the sampled Gaussian noise, and $v_Z$ the predictor's velocity under the tactile-augmented condition. We supervise $v_Z$ with the conditional flow-matching target $\varepsilon-y$. Although the implementation expresses this objective relative to the unaugmented prediction, the resulting loss is algebraically identical to the conditional flow-matching loss. The predictor's parameters are frozen, but gradients through its tactile-augmented forward pass update the adapter and tactile encoder. The squared error is spatially weighted, normalized by the weighted valid element count for each sample--horizon pair, and averaged over valid pairs. Reference ROI weights are 1 for the hand region and 0.1 for the background.

\textbf{Future-pressure prediction.}
The pressure predictor uses the tactile context, detached cached visual memory, and a detached window of upcoming edited actions. Detaching the action window prevents this loss from sending gradients through the action condition into the residual readout, while gradients through the tactile context can update the shared tactile encoder. The target is measured future pressure, not a learned tactile embedding. For hand $r\in\{\mathrm L,\mathrm R\}$ and pressure site $u\in\{1,\ldots,441\}$, a fixed codec canonicalizes hand orientation and standardizes the future pressure $P^{+,r}_u$ using the corresponding training-set mean $\mu^r_u$ and standard deviation $\sigma^r_u$:
\begin{equation}
 y^r_u=\frac{P^{+,r}_u-\mu^r_u}
                  {\max(\sigma^r_u,10^{-3})}.
 \label{eq:app-pressure-codec}
\end{equation}
Future-hand validity and spatial sensor masks select the sites included in the flow-matching loss, which is averaged over all valid sites rather than weighting the two hands equally. In the contact-filtered variant, the future-hand mask also includes the gate computed from current tactile history; no gate is computed from the future target. The reference predictor uses four Transformer decoder layers of width 256 with eight attention heads.

\subsection{Training Strategy}
\label{app:training-stages}
For tactile pretraining, synchronized tactile and hand-pose histories from EgoTouch are encoded into normalized embeddings $u_i$ and $p_i$, with tactile embeddings taken before reference subtraction.
Their similarities are computed as $S_{ij}=\alpha u_i^\top p_j$, where $\alpha>0$ is a learned scale.
The symmetric contrastive objective averages row-wise cross-entropy over $S$ and $S^\top$, treating synchronized pairs as positives and other pairs in the minibatch as negatives.
During editor training, the visual branch supplies cached visual memory and base action chunks and remains frozen.
The tactile encoder, residual editor, tactile conditioning adapter, and pressure predictor are updated, while the visual predictor remains frozen.
The additional experiments retain a frozen visual backbone during editor training. Joint fine-tuning of the two control branches is a possible extension, not a stage evaluated in these experiments.

\section{Action and Evaluation Conventions}
\label{app:protocol}

\subsection{Action Layout and Temporal Alignment}
\label{app:action-interface}

\textbf{Action coordinates.}
The hand-motion interface uses $H=16$ action positions and 192 channels per position. Of these, 138 channels contain motion data and 54 are padding. Table~\ref{tab:app-action-layout} gives the component groups used by the losses. Padding is excluded from supervision and from valid sampled action components.

\textbf{Chunk positions and auxiliary action windows.}
Each visual update produces a new base action chunk. Between visual updates, the editor revises its remaining actions using the latest available tactile history. Offline replay uses a nominal 10~Hz visual schedule and 10 or 30~Hz editing, whereas hardware timing is measured separately in Appendix~\ref{app:hardware-timing}. Elapsed time is referenced to the active chunk. The additional Tachin reproduction retains zero elapsed-time conditioning inside the editor, while actual decision time still determines its editable-action mask.

For future-pressure conditioning, the action window begins at $k_0=\lceil f_{\mathrm{cmd}}\Delta t_{t,\tau}\rceil$, where $f_{\mathrm{cmd}}$ is the action command rate, and contains the next eight edited actions. A floating-point tolerance is applied at integer boundaries. Positions beyond the chunk are zero-padded and marked invalid. This auxiliary window-indexing rule is distinct from the loss horizon and from the temporal mask that marks already executed actions as non-editable.

\subsection{Evaluation Masks and Aggregation}
\label{app:evaluation}

\textbf{Physical hand-motion errors.}
The hand-motion evaluator supports a declared horizon $H_{\mathrm{eval}}$ of eight or sixteen positions. The reference evaluator requires a complete valid prefix. The additional cached-replay experiments use their archived masks, including per-position validity for EgoTouch; their results are reported separately rather than treated as interchangeable with this reference protocol. Predictions and targets are compared in the physical action coordinates used by the evaluator.

\input{table/tab-4}

At prediction step $k$, wrist translation increments are accumulated from the start of the prediction window before computing the Euclidean error in centimeters. Euler rotation increments are converted using the $xyz$ convention and composed by left multiplication, $R_k=\delta R_kR_{k-1}$ with $R_0=I$; predicted rotations are composed in the same way. The resulting rotational error, in degrees, is
\begin{equation}
 e_{\mathrm{rot}}(R,\widehat R)
 =\frac{180}{\pi}\arccos\!\left(
   \operatorname{clip}_{[-1,1]}\!\left(
   \frac{\operatorname{tr}(\widehat R^{\!\top}R)-1}{2}
   \right)
 \right).
 \label{eq:app-rotation-metric}
\end{equation}
Finger-joint MPJPE compares the stored palm-local positions directly, averaging joints 1--20 and reporting centimeters. Joint 0 is the fixed local origin and is excluded; the evaluator does not subtract a predicted wrist position from already local joint coordinates.

\textbf{Sequence-level aggregation.}
For each metric, errors are first summed over eligible hand--time entries within each sequence and divided by their valid count. Sequences with no eligible entries are excluded. Environment-level means then weight these sequence means equally. The EgoTouch macro-average gives equal weight to its five environment means; the Tachin overall score gives equal weight to its eligible sequence means. This aggregation differs from pooling all frames and should not be replaced by a sample-weighted mean from a different evaluation entry point.

\textbf{Contact-conditioned evaluation.}
In the contact-filtered protocol, a sample is admitted when at least one supervised hand satisfies the causal gate in Eq.~\ref{eq:app-contact-gate}. Per-hand masks then exclude noncontact hands from action and future-pressure supervision. These exclusions define a contact-conditioned subset of the held-out data, not the complete held-out
test set. Retaining the base action for an excluded hand does not insert that hand into the metric denominator. Comparisons under this protocol require a common sample set, hand mask, and prediction horizon; results on an unfiltered test set represent a separate evaluation protocol.

\section{Visual-Only Action Prediction}
\label{app:visual-only}

The visual-only experiment evaluates future hand-motion prediction when touch is unavailable. Tactile observations are excluded during both training and evaluation, and tactile supervision is disabled during training. The model uses only RGB, language, pose history, recurrent visual memory, and the visual action expert. Thus,
\begin{equation}
A_t=A_t^{\mathrm{base}},\qquad
 \mathcal L_{\mathrm{visual\text{-}only}}
 =\mathcal L_{\act}+\lambda_{\vis}\mathcal L_{\vis}.
 \label{eq:no-touch}
\end{equation}
This setting isolates the visual pathway rather than testing cross-modal transfer.

Table~\ref{tab:visual-only-prediction} shows that \method{} remains competitive without touch and achieves the lowest error on eight of the nine dataset-level metric comparisons. On GigaHands, \method{} obtains average translation, rotation, and joint errors of $0.65$~cm, $3.64^\circ$, and $0.43$~cm. Relative to SmolVLA, these values correspond to reductions of $25.3\%$, $15.9\%$, and $54.7\%$, respectively. The improvement is consistent across all five activity families rather than being concentrated in a single category.

On OakInk2, \method{} reaches $0.44$~cm translation error, $1.38^\circ$ rotation error, and $0.37$~cm joint MPJPE, reducing the corresponding SmolVLA errors by $12.0\%$, $32.0\%$, and $67.3\%$. On TACO, \method{} achieves the lowest rotation and joint errors at $3.40^\circ$ and $0.46$~cm, improving over SmolVLA by $7.6\%$ and $55.3\%$. Its translation error is $0.75$~cm, slightly above SmolVLA's $0.72$~cm but still $22.7\%$ below VITRA. These results show that separating tactile correction from visual action generation does not weaken the visual predictor; instead, the visual memory and predictive supervision provide a strong action prior even when the tactile branch is absent.

\input{table/tab-5}

\section{Effects of Tactile Input and Update Timing}
\label{app:tactile-input}

We measure how much of the full tactile editor's gain is retained
when changing the number and timing of edits or the tactile input.
Full \method{} performs three tactile edits per visual update.
We compare it with one edit using sparsely sampled tactile history,
and with two edits at either 0 and 33\,ms or 0 and 67\,ms.
All times are measured from the start of the visual action chunk.
We also evaluate two input conditions: holding the tactile input
fixed throughout the chunk, and reversing the contact-strength
signal to produce mismatched touch.

For each error metric $m$, let $E_{c,m}$ denote the error under
condition $c$, $E_{\mathrm{full},m}$ the error of full \method{},
and $E_{\mathrm{off},m}$ the error with tactile editing disabled.
When the full editor reduces error, we report the fraction of
its gain lost under condition $c$:
\begin{equation}
  \mathrm{GainLoss}_{c,m}
  = \frac{E_{c,m}-E_{\mathrm{full},m}}
         {E_{\mathrm{off},m}-E_{\mathrm{full},m}}
    \times 100\%.
  \label{eq:editing-gain-loss}
\end{equation}
A value of $0\%$ means that the condition retains the full
editing gain, and $100\%$ corresponds to the editor-off error.
Negative values indicate lower error than full \method{}.
For example, the one-edit condition with sparse tactile history
loses $3.1\%$ of the finger-joint editing gain:
$(0.2613-0.2563)/(0.4168-0.2563)\times 100\%$.

Table~\ref{tab:tactile-editing-gain} shows that both two-edit
schedules lose only $0.9\%$ of the finger-joint gain.
Moving the second edit from 33 to 67\,ms reduces gain loss
for both wrist metrics.
Holding tactile input fixed changes gain loss by at most
$0.3$ percentage points relative to the full editor.
Mismatched touch produces the largest observed losses,
including $9.2\%$ of the wrist-translation gain and $7.2\%$
of the finger-joint gain.
Under this protocol, the editor is more sensitive to tactile
mismatch than to the tested changes in update timing.

\section{Real-Robot Setup and Demonstration Collection}
\label{app:robot-setup}

\textbf{Hardware.}
Our platform is a dual-arm Linker A7black robot with a 7-DoF arm and a Linker Hand L6 dexterous hand on each side. The hands provide fingertip tactile sensing. We mounted an Intel RealSense D435i on a custom 3D-printed bracket to capture an egocentric view of the workspace and calibrated its pose relative to the robot. 

\begin{figure}[htbp]
    \centering
\vspace{-0.4cm}
    \begin{minipage}[t]{0.47\linewidth}
        \centering
\includegraphics[width=\linewidth,height=4.2cm,keepaspectratio]
        {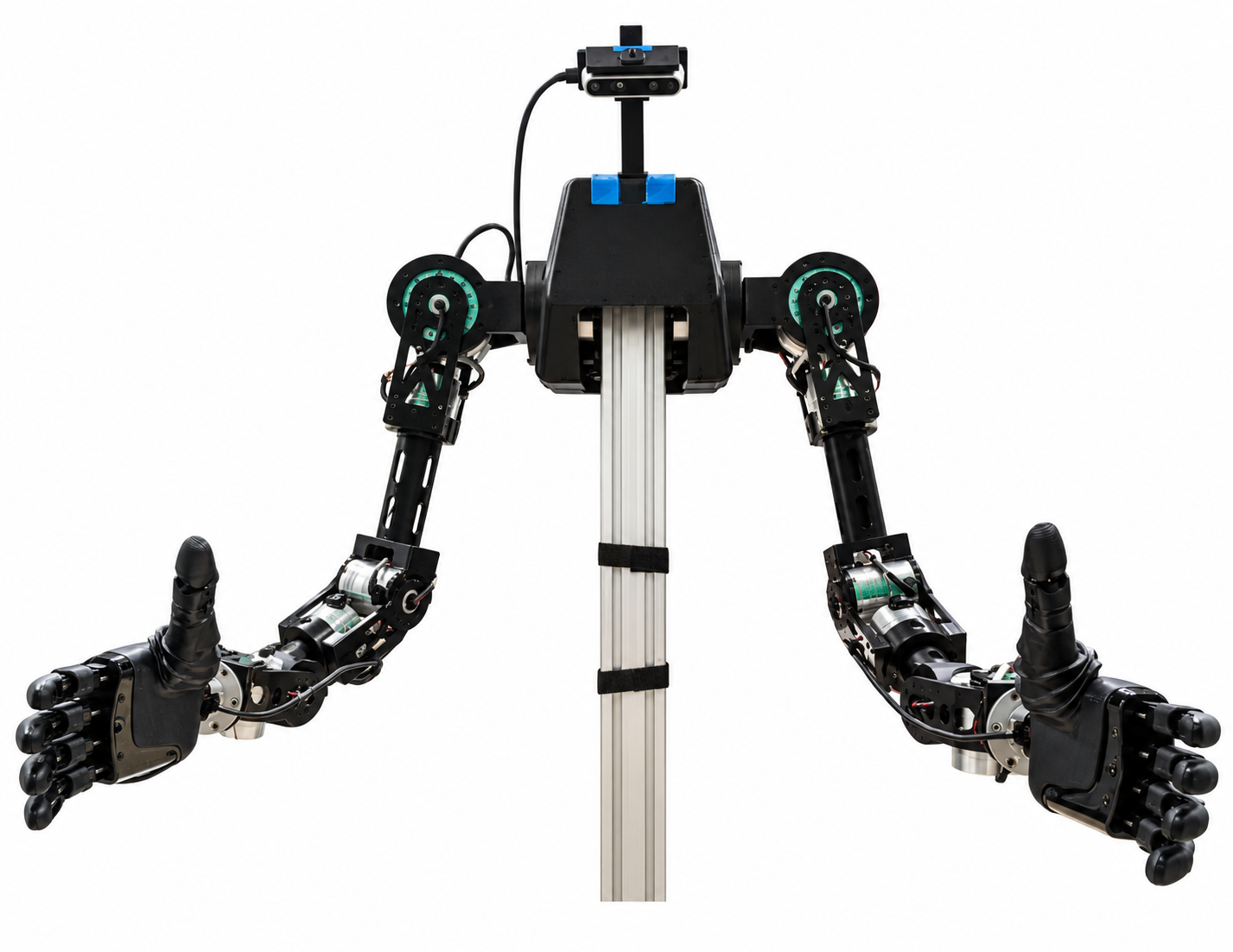}
        \vspace{2pt}
        \parbox{\linewidth}{\centering\small
        (a) Robot hardware platform}
    \end{minipage}
    \hfill
    \begin{minipage}[t]{0.47\linewidth}
        \centering
\includegraphics[width=\linewidth,height=4.2cm,keepaspectratio]
        {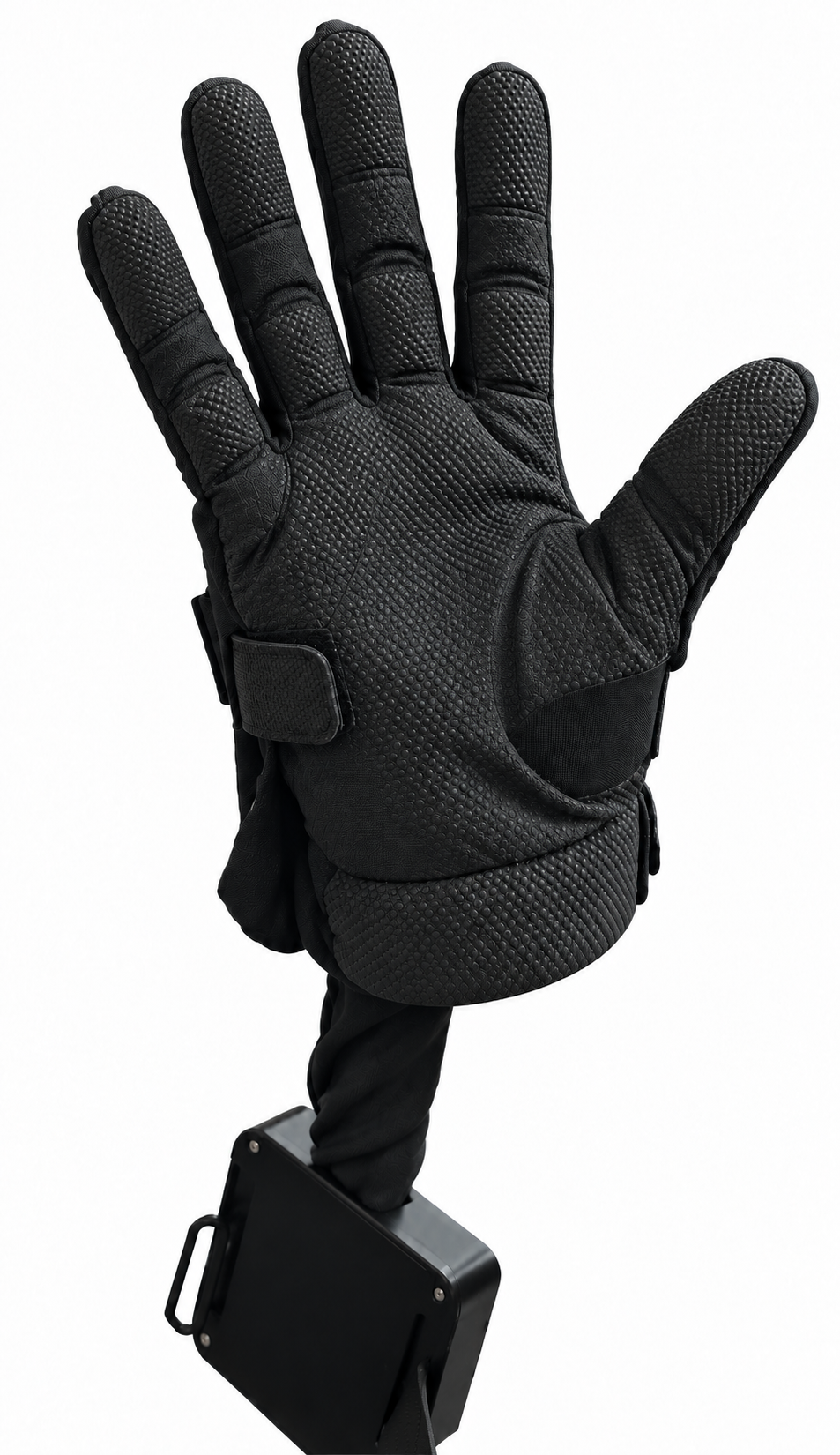}
        \vspace{2pt}
        \parbox{\linewidth}{\centering\small
        (b) Data glove}
    \end{minipage}
    \vspace{-0.4cm}
    \caption{\textbf{Robot platform and data glove}}
    \label{fig:hardware_and_glove}
    \vspace{-0.4cm}
\end{figure}

\textbf{Teleoperation.}
An operator wears Wuji data gloves to capture finger articulation and HTC VIVE Tracker 3.0 devices at the wrists to track wrist motion. Two SteamVR Base Station 2.0 units track the VIVE devices. The tracked wrist poses and glove measurements are mapped to the robot arms and dexterous hands to collect task demonstrations; no VR headset is required.

\textbf{Demonstration budget.}
We collect 30 teleoperated episodes for each of the four real-robot tasks: egg transfer, single-sheet pickup, paper-cup nesting, and paper-bag dragging.

\section{Additional Qualitative Results}
\label{app:appendix_qualitative}

We provide additional qualitative comparisons on EgoTouch and Tachin to complement the examples shown in Fig.~\ref{fig:qualitative_main}.
Throughout this section, we use the same configurations as in Table~\ref{tab:tactile-prediction}. VITRA, SmolVLA, and MemoryVLA use their tactile-augmented variants. T-Rex and \method{} use their native tactile interfaces.
We follow the same visualization protocol as Fig.~\ref{fig:qualitative_main}. For each example, the source RGB observation is shown at $t_0$, while the ground-truth pose and all model predictions correspond to $t_0 + 100\,\mathrm{ms}$.
Camera-frame overlays visualize the complete predicted hand pose, whereas wrist-aligned views emphasize differences in finger articulation.
The value below each prediction is the per-example wrist-aligned MPJPE over the 20 non-root hand joints at $t_0 + 100\,\mathrm{ms}$.
These values are intended only as diagnostics for the displayed examples and should not be interpreted as dataset-level test averages.
They isolate finger-articulation error after wrist alignment and are therefore distinct from the wrist-translation and wrist-rotation metrics reported in Table~\ref{tab:tactile-prediction}.
Across the displayed examples, \method{} consistently produces fingertip configurations that more closely match the ground truth and achieves the lowest per-example wrist-aligned MPJPE among the compared methods.
This mirrors the quantitative results in Table~\ref{tab:tactile-prediction}: the largest gains from \method{} occur in finger-joint MPJPE, substantially exceeding the relative improvements in wrist translation and rotation.
Together, these results suggest that \method{} is particularly effective at predicting fine-grained fingertip and finger-articulation dynamics.

\subsection{Additional EgoTouch Qualitative Examples}

\begin{figure*}[!t]
    \centering
    \includegraphics[width=\linewidth,height=.83\textheight,keepaspectratio]{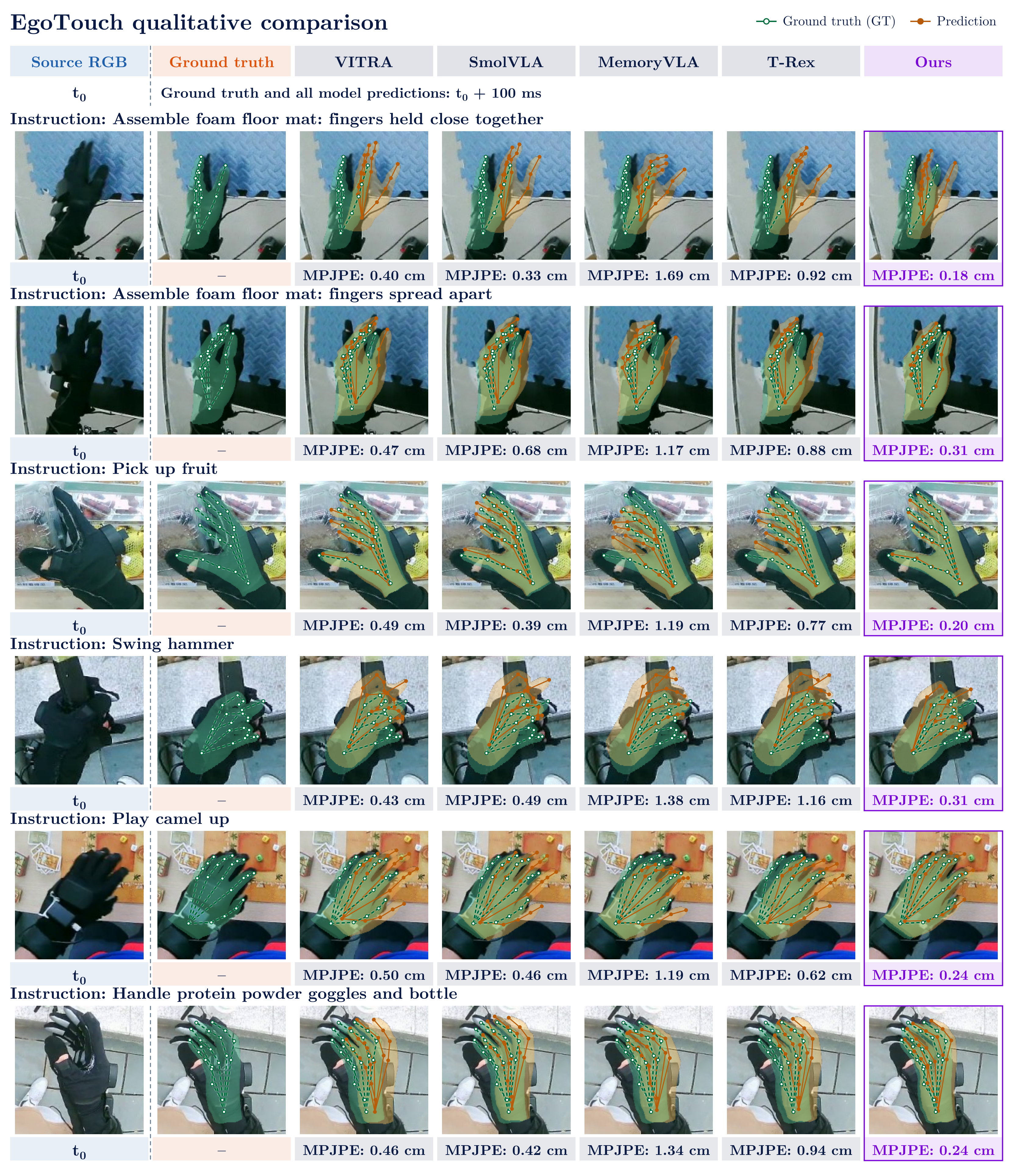}
    \caption{
    \textbf{Additional camera-frame qualitative comparisons on EgoTouch.}
    Each row shows the source RGB observation at $t_0$, followed by the ground-truth hand pose
    and predictions from VITRA, SmolVLA, MemoryVLA, T-Rex, and \method{} at
    $t_0 + 100\,\mathrm{ms}$.
    Ground truth is shown in green and model predictions in orange.
    Values below each prediction report the corresponding wrist-aligned finger-joint MPJPE
    in centimeters.
    }
    \label{fig:appendix_egotouch_rgb}
\end{figure*}

\begin{figure*}[!t]
    \centering
    \includegraphics[width=\linewidth,height=.83\textheight,keepaspectratio]{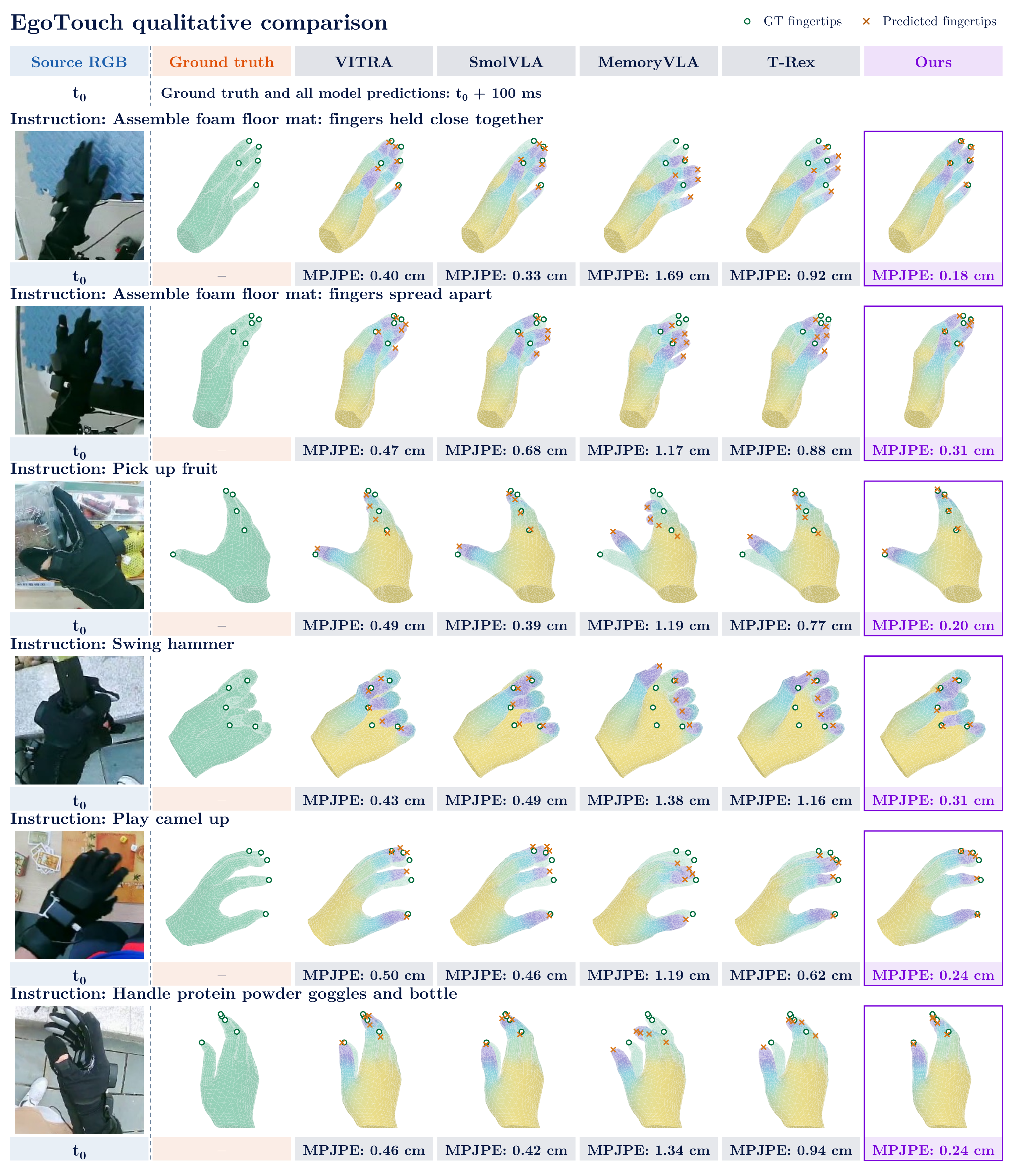}
    \caption{
    \textbf{Additional wrist-aligned fingertip comparisons on EgoTouch.}
    The same six examples from Fig.~\ref{fig:appendix_egotouch_rgb} are shown after wrist alignment
    to emphasize finger articulation independently of global wrist motion.
    Within each example, all hands share the same alignment, scale, and
    $+60^\circ$ visualization rotation.
    The ground-truth hand mesh is shown in green, with ground-truth fingertips marked by green circles
    and predicted fingertips marked by orange crosses.
    Values below each prediction report wrist-aligned MPJPE over the 20 non-root hand joints at
    $t_0 + 100\,\mathrm{ms}$.
    These values are per-example diagnostics rather than dataset-level test averages.
    }
    \label{fig:appendix_egotouch_mesh}
\end{figure*}

Figures~\ref{fig:appendix_egotouch_rgb} and~\ref{fig:appendix_egotouch_mesh}
show six additional EgoTouch examples.
We first visualize predictions in the camera frame and then use wrist-aligned renderings
to isolate differences in finger articulation.

\subsection{Additional Tachin Qualitative Examples}

\begin{figure*}[!t]
    \centering
    \includegraphics[width=\linewidth,height=.83\textheight,keepaspectratio]{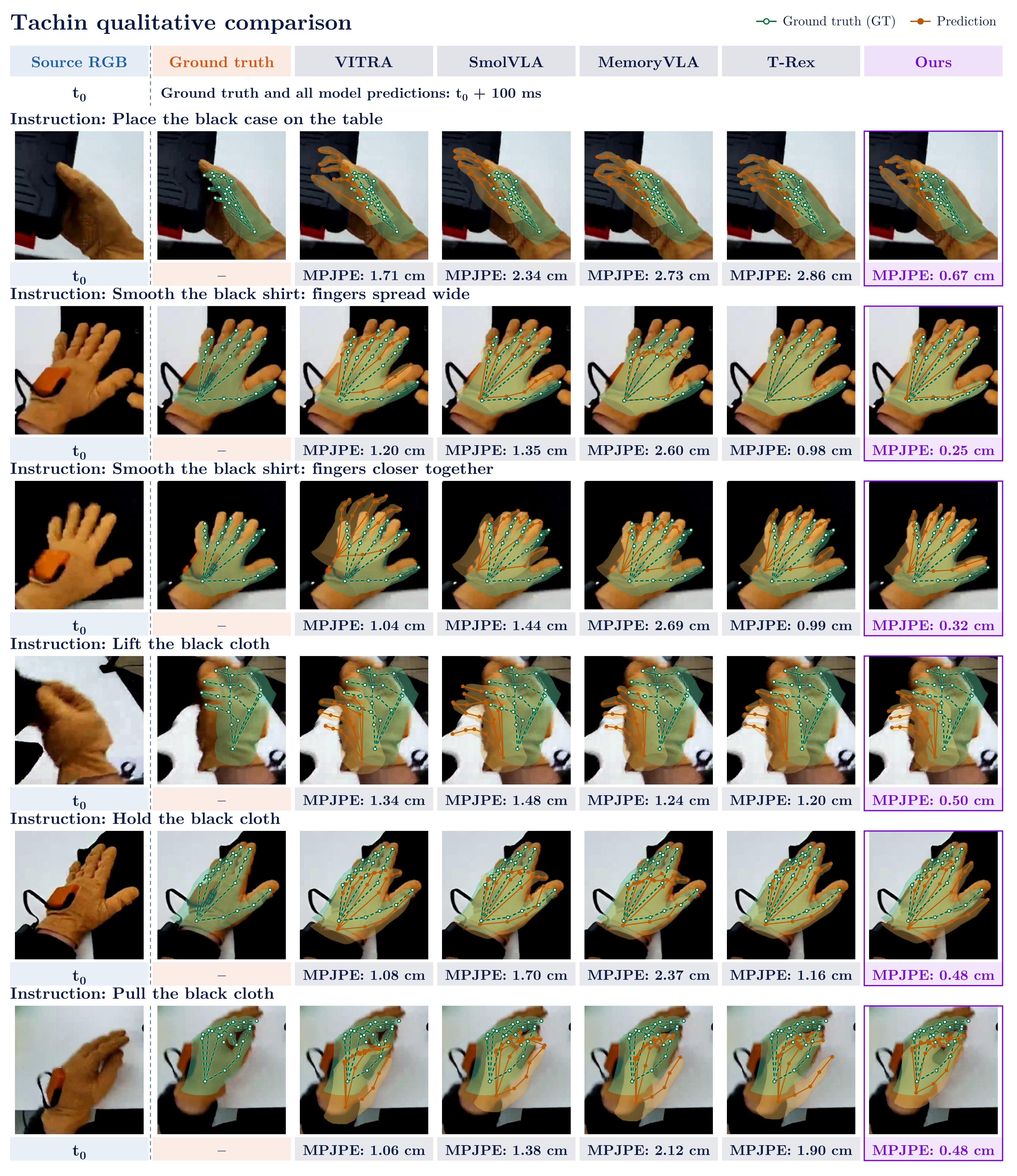}
    \caption{
    \textbf{Additional camera-frame qualitative comparisons on Tachin.}
    Each row shows one source RGB observation at $t_0$, followed by the ground-truth hand pose and predictions from VITRA, SmolVLA, MemoryVLA, T-Rex, and \method{} at $t_0 + 100\,\mathrm{ms}$.
    Ground-truth hand poses are shown in green and model predictions in orange.
    This camera-frame visualization preserves both global wrist motion and finger articulation.
    The values below each prediction report the corresponding wrist-aligned finger-joint MPJPE in centimeters.
    }
    \label{fig:appendix_tachin_rgb}
\end{figure*}

\begin{figure*}[!t]
    \centering
    \includegraphics[width=\linewidth,height=.79\textheight,keepaspectratio]{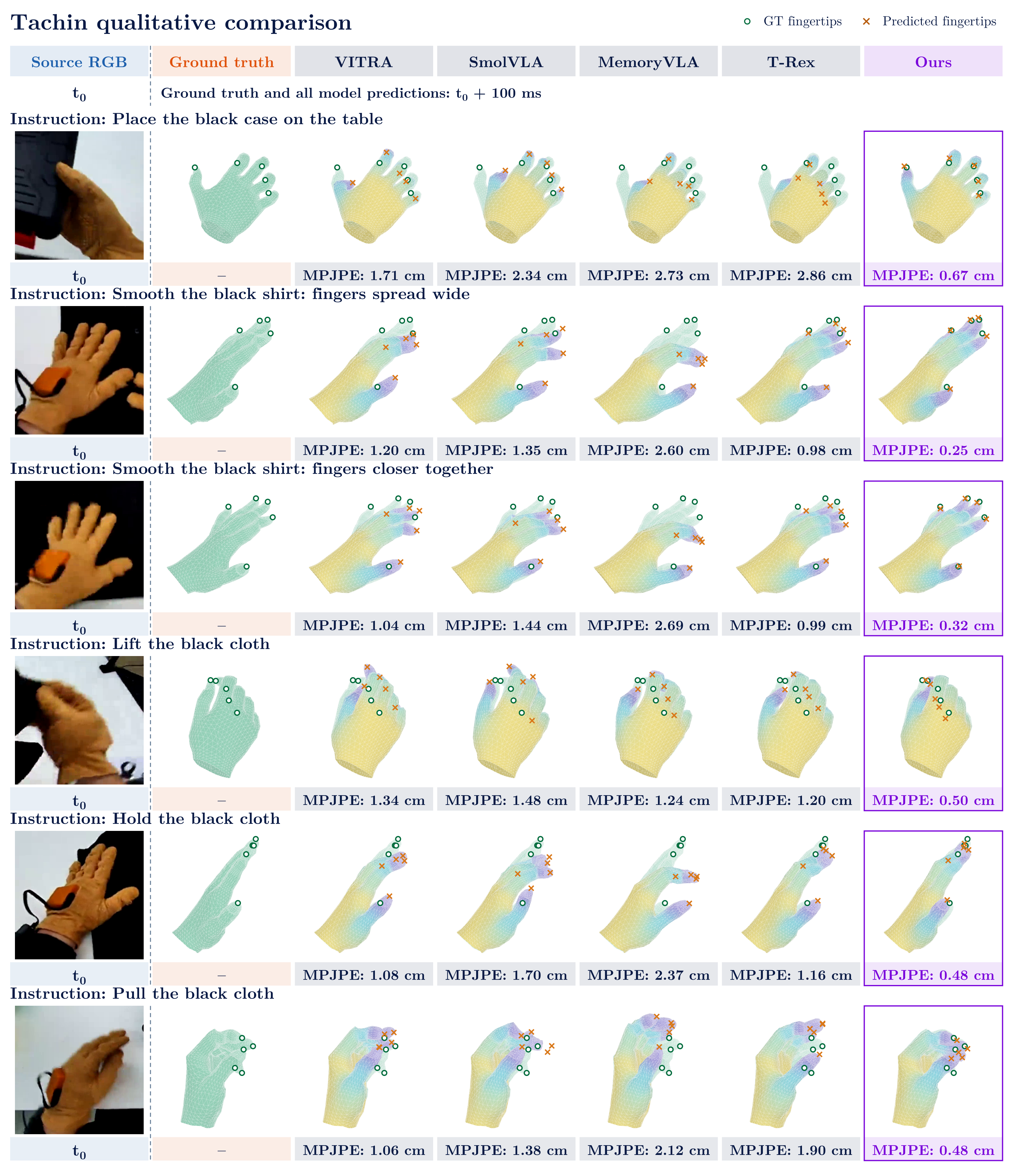}
    \caption{
    \textbf{Additional wrist-aligned fingertip comparisons on Tachin.}
    The same six examples from Fig.~\ref{fig:appendix_tachin_rgb} are shown after wrist alignment to emphasize finger articulation independently of global wrist motion.
    Within each example, all hands share the same alignment, scale, and $+60^\circ$ visualization rotation.
    The ground-truth hand mesh is shown in green, with ground-truth fingertips marked by green circles and predicted fingertips marked by orange crosses.
    Values below each prediction report wrist-aligned MPJPE over the 20 non-root hand joints at $t_0 + 100\,\mathrm{ms}$.
    These values are per-example diagnostics rather than dataset-level test averages.
    }
    \label{fig:appendix_tachin_mesh}
\end{figure*}

Figures~\ref{fig:appendix_tachin_rgb} and~\ref{fig:appendix_tachin_mesh}
show six additional Tachin examples covering diverse hand orientations and finger configurations.
We first visualize the future hand-pose predictions directly in the camera frame, where the source RGB observation is shown at $t_0$ and the ground-truth and predicted poses correspond to $t_0 + 100\,\mathrm{ms}$.
This view preserves both global wrist motion and finger articulation, allowing direct comparison of the complete predicted hand pose.

We then show wrist-aligned fingertip-detail renderings for the same examples.
Removing global wrist displacement makes differences in finger articulation easier to inspect independently of the hand's position in the camera frame.
Across the displayed examples, \method{} more closely matches the ground-truth fingertip configuration and achieves the lowest per-example wrist-aligned MPJPE in every row.
These qualitative examples complement the dataset-level quantitative results reported in Table~\ref{tab:tactile-prediction}.

\subsection{Additional Real-Robot Qualitative Examples}
\label{app:realrobot-qualitative}

\begin{figure*}[!t]
    \centering
    \includegraphics[width=\linewidth,height=.83\textheight,keepaspectratio]{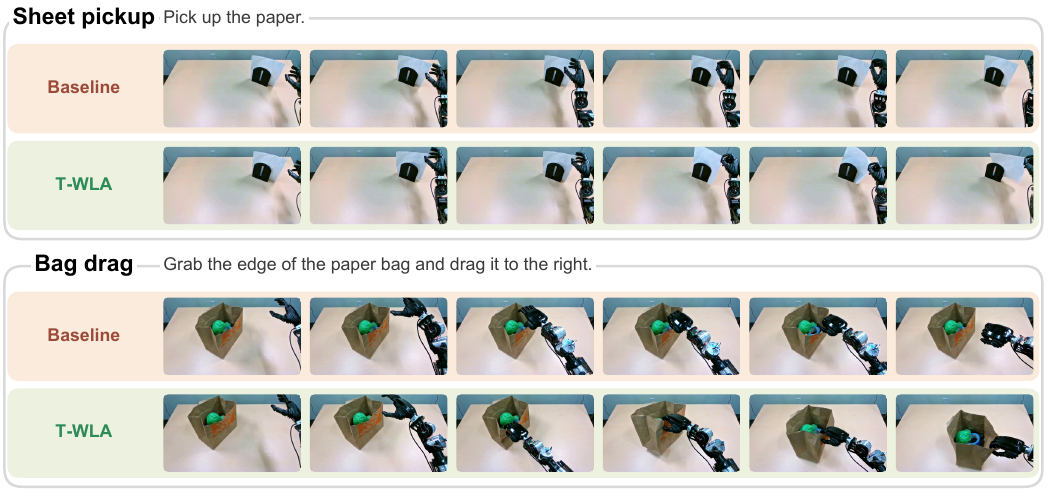}
    \caption{
    \textbf{Additional real-robot qualitative examples.}
    Selected frames from paper-bag dragging and single-sheet pickup.
    Both tasks involve grasping thin, deformable edges, requiring sustained contact during dragging or a delicate pinch during lifting.
    }
    \label{fig:appendix_robot_bag_sheet}
\end{figure*}

Figure~\ref{fig:appendix_robot_bag_sheet} presents additional real-robot examples of paper-bag dragging and single-sheet pickup.
Both tasks require precise contact with a thin edge, but impose different demands after grasping: sheet pickup requires maintaining a secure pinch without crumpling the paper, whereas dragging a loaded bag requires sustaining the grasp under changing fingertip loads.
Tactile feedback provides contact information relevant to these adjustments, complementing visual guidance.
These examples supplement the closed-loop success rates reported in Table~\ref{tab:robot-success}.

\section{Hardware-in-the-Loop Command-Pipeline Timing}
\label{app:hardware-timing}

We additionally report a hardware-in-the-loop profile on an RTX 4070 Laptop GPU with 8\,GB memory, using a half-precision, compiled visual model.
The pipeline includes camera capture, visual prediction, tactile editing, inverse kinematics (IK), and command dispatch.
The camera, Linker L6 right hand, and its five $12\times6$ fingertip sensors are physical devices; hand communication uses a 1\,Mbps CAN bus.
The arm is replaced by an ideal target-following substitute and does not physically move.
Thus, this profile measures command-pipeline operation, not complete closed-loop execution by a moving robot arm.
Each eight-step chunk schedules three edits, before steps 0, 3, and 6, modifying only actions that have not yet been dispatched.

\begin{table}[!t]
\centering
\caption{Hardware-in-the-loop throughput. The entire reported run is the primary aggregate; the IK-converged subset is a conditional measurement from the same run. Editor rates count scheduled correction outputs, not independent fresh full-hand observations or confirmed physical execution. Durations and rates are rounded as reported.}
\label{tab:hardware-timing}
\small
\setlength{\tabcolsep}{3pt}
\begin{tabular}{@{}lrrrr@{}}
\toprule
Scope & Cycles & Time (s) & \shortstack{Visual\\(Hz)} & \shortstack{Editor\\(Hz)} \\
\midrule
Entire run & 395 & 119 & 3.31 & 9.94 \\
IK-converged subset & 50 & 10.3 & 4.84 & 14.5 \\
\bottomrule
\end{tabular}
\end{table}

Table~\ref{tab:hardware-timing} distinguishes the full-run rates from the selected 50 cycles with converged IK, for which three warm-up cycles were excluded.
Most cycles in the complete run reached the IK iteration limit.
The approximately threefold editor-to-visual rate ratio follows the three-edits-per-chunk schedule and is not a measurement of standalone editor throughput.

\textbf{Selected-window sensor and command rates.}
The remaining reported statistics apply to the selected IK-converged window.
Tactile reception is $8.3$\,Hz per fingertip, with $41.4$ individual-finger frames received per second across all five fingers.
The latter is an aggregate count, not a $41.4$\,Hz full-hand observation stream; sequential finger reads also introduce temporal skew.
The $14.5$\,Hz correction outputs are reported to enter outgoing commands, but fresh complete tactile input at every edit is not established.
The within-chunk action-dispatch setting is $80$\,Hz, while the measured hand-command rate over this window is $38.7$ commands/s.
Hand commands and tactile reads share the same bus, so their rates depend on the dispatch workload.

\textbf{Latency endpoints.}
Host frame-receipt-to-command-send latency has P50/P95 values of $36/57$\,ms for hand commands and $28/48$\,ms for arm-target commands.
All timestamps use the control computer's clock, and the start timestamp is completion of tactile-frame reception on that host.
The L6 provides no hardware sampling timestamp, and neither controller acknowledges command application.
These measurements therefore exclude an unknown physical-sampling-to-host interval and do not measure controller-application or actuation latency.

\newpage

%% file: table/tab-3-1.tex
\begin{table*}[t]
\centering
\caption{\textbf{Comparison with tactile-based methods.}
Fast touch: tactile refinement between visual replans.
V/T: vision/touch; Aux.: auxiliary prediction objective;
Online: predicted sensory output used at inference.
$^{*}$Planner-level context; --: not applicable or not specified.}
\vspace{-0.4cm}
\label{tab:related-work-comparison}
\begingroup
\small
\setlength{\tabcolsep}{3pt}
\renewcommand{\arraystretch}{1.12}
\begin{tabular*}{\linewidth}{@{\extracolsep{\fill}}llclll@{}}
\toprule
\textbf{Method} & \textbf{Visual context} & \textbf{Fast touch} & \textbf{Tactile mechanism} & \textbf{Prediction} & \textbf{Use}\\
\midrule
RDP~\citep{xue2025reactive}        & RGB window     & Yes & Latent decoder   & --          & --\\
T-Rex~\citep{niu2026t}            & Chunk cache    & Yes & Flow refinement  & Future V    & Aux.\\
ViTaR~\citep{wang2026vitar}        & Current RGB    & No  & Action residual  & --          & --\\
ResTacVLA~\citep{zhang2026feeling}  & Current RGB    & --  & Feature residual & Current T   & Online\\
TouchWorld~\citep{zhou2026touchworld} & Task memory$^{*}$ & Yes & Action residual & Future V+T & Online\\
ViTaL~\citep{wu2026inference}      & Latent window  & --  & Diffusion edit   & Future V+T  & Online\\
\midrule
\textbf{\method{} (ours)} & \textbf{Recurrent bank} & \textbf{Yes} & \textbf{Action residual} & \textbf{Future V+T} & \textbf{Aux.}\\
\bottomrule
\end{tabular*}
\endgroup
\vspace{-0.6cm}
\end{table*}

%% file: table/tab-4.tex
\begin{table*}[!t]
\centering
\caption{Visual-only future hand-motion prediction. Trans., Rot., and
MPJPE give wrist translation error (cm), wrist rotation geodesic error
($^\circ$), and wrist-relative finger-joint MPJPE (cm), respectively.
GigaHands Avg. weights the five activity families equally. OakInk2 and
TACO use their complete held-out test sets.
\textbf{Bold} and \underline{underline} mark the best and second-best
distinct errors within each row and metric, including ties. Gray
highlights \method{}. Frequencies are shown in Hz.}
\vspace{-0.3cm}
\label{tab:visual-only-prediction}
\footnotesize
\setlength{\tabcolsep}{3pt}
\renewcommand{\arraystretch}{1.12}
\resizebox{\linewidth}{!}{%
\begin{tabular}{@{}ll*{9}{r}*{3}{>{\columncolor[gray]{0.93}}r}@{}}
\toprule
& & \multicolumn{3}{c}{VITRA}
  & \multicolumn{3}{c}{SmolVLA}
  & \multicolumn{3}{c}{MemoryVLA}
  & \multicolumn{3}{c@{}}{\cellcolor[gray]{0.93}\textbf{\method{}}} \\
\cmidrule(lr){3-5}\cmidrule(lr){6-8}
\cmidrule(lr){9-11}\cmidrule(l){12-14}
Dataset & Activity
& Trans. & Rot. & MPJPE
& Trans. & Rot. & MPJPE
& Trans. & Rot. & MPJPE
& Trans. & Rot. & MPJPE \\
\midrule
GigaHands & Cook.
& 1.16 & 5.32 & 1.48
& \underline{0.84} & \underline{3.75} & \underline{0.96}
& 1.32 & 6.52 & 4.00
& \textbf{0.63} & \textbf{3.19} & \textbf{0.43} \\
& Office
& 1.25 & 5.79 & 1.32
& \underline{0.81} & \underline{3.91} & \underline{0.94}
& 1.26 & 6.54 & 3.91
& \textbf{0.60} & \textbf{3.41} & \textbf{0.43} \\
& Craft.
& 1.11 & 5.13 & 1.36
& \underline{0.73} & \underline{3.69} & \underline{0.94}
& 1.23 & 6.07 & 3.78
& \textbf{0.56} & \textbf{3.13} & \textbf{0.41} \\
& Entert.
& 1.44 & 6.72 & 1.41
& \underline{1.01} & \underline{5.07} & \underline{0.95}
& 1.35 & 6.97 & 3.90
& \textbf{0.69} & \textbf{4.01} & \textbf{0.44} \\
& Housew.
& 1.37 & 7.00 & 1.36
& \underline{0.98} & \underline{5.23} & \underline{0.95}
& 1.35 & 7.25 & 3.99
& \textbf{0.74} & \textbf{4.45} & \textbf{0.45} \\
\addlinespace[2pt]
& \textbf{Avg.}
& 1.27 & 5.99 & 1.39
& \underline{0.87} & \underline{4.33} & \underline{0.95}
& 1.30 & 6.67 & 3.92
& \textbf{0.65} & \textbf{3.64} & \textbf{0.43} \\
\midrule
OakInk2 & Overall
& 0.53 & \underline{2.03} & 2.04
& \underline{0.50} & \underline{2.03} & \underline{1.13}
& 1.05 & 5.27 & 3.92
& \textbf{0.44} & \textbf{1.38} & \textbf{0.37} \\
TACO & Overall
& 0.97 & 4.76 & 2.74
& \textbf{0.72} & \underline{3.68} & \underline{1.03}
& 1.20 & 6.66 & 4.02
& \underline{0.75} & \textbf{3.40} & \textbf{0.46} \\
\midrule
\multicolumn{2}{@{}l}{Frequency (Hz)}
& \multicolumn{3}{c}{19.4}
& \multicolumn{3}{c}{23.0}
& \multicolumn{3}{c}{8.8}
& \multicolumn{3}{c@{}}{\cellcolor[gray]{0.93}39.7} \\
\bottomrule
\end{tabular}%
}
\vspace{-0.6cm}
\end{table*}

%% file: table/tab-5.tex
\begin{table}[t]
\centering
\caption{Editing gain lost under different tactile inputs and
update schedules. Edit counts are per visual update, and times
are measured from the start of the visual action chunk.
Trans., Rot., and Finger refer to wrist translation, wrist
rotation, and wrist-relative finger-joint MPJPE, respectively.
Lower gain loss is better.}
\label{tab:tactile-editing-gain}
\small
\setlength{\tabcolsep}{3pt}
\begin{tabular}{@{}lrrr@{}}
\toprule
Condition & Trans. & Rot. & Finger \\
\midrule
One edit, sparse history
    & 3.7\% & 3.0\% & 3.1\% \\
Two edits: 0, 33\,ms
    & 3.1\% & 3.1\% & 0.9\% \\
Two edits: 0, 67\,ms
    & 1.2\% & 1.3\% & 0.9\% \\
\textbf{Full \method{}} (three edits)
    & 0.0\% & 0.0\% & 0.0\% \\
Fixed tactile input
    & $-0.3$\% & 0.1\% & 0.2\% \\
\shortstack[l]{Mismatched touch\\
(reversed contact strength)}
    & 9.2\% & 3.3\% & 7.2\% \\
\bottomrule
\end{tabular}
\end{table}